\documentclass[acmtog]{acmart}

\AtBeginDocument{%
  \providecommand\BibTeX{{%
    \normalfont B\kern-0.5em{\scshape i\kern-0.25em b}\kern-0.8em\TeX}}}

\setcopyright{acmcopyright}\acmJournal{TOG}
\acmYear{2021}\acmVolume{40}\acmNumber{6}\acmArticle{213}\acmMonth{12} \acmDOI{10.1145/3478513.3480485}

\usepackage{multirow}
\usepackage{booktabs}
\usepackage{array}
\usepackage{float}
\usepackage{appendix}
\usepackage{epsfig}
\usepackage{graphicx}
\usepackage{amsmath}
\usepackage{tabularx}
\usepackage{contour}
\usepackage[normalem]{ulem}

\contourlength{0.8pt}

\newcommand{\rurl}[1]{%
  \textcolor{magenta}{
  \uline{\phantom{#1}}%
  \llap{\contour{white}{#1}}}
}

\newcolumntype{P}[1]{>{\centering\arraybackslash}m{#1}}

\newcommand{\xuan}[1]{#1}
\newcommand{\update}[1]{#1}

\newcommand{\mypar}[1]{\paragraph{#1}}

\newcommand{\stylegantwo}[1]{{StyleGAN2}}

\begin{document}

\title{Time-Travel Rephotography}

\author{Xuan Luo}
\email{xuanluo@cs.washington.edu}
\orcid{1234-5678-9012}
\affiliation{%
  \institution{University of Washington}
}

\author{Xuaner (Cecilia) Zhang}
\affiliation{%
  \institution{Adobe Inc.}
}
\email{cecilia77@berkeley.edu}

\author{Paul Yoo}
\affiliation{%
  \institution{University of Washington}
}
\email{yoosehy@cs.washington.edu}

\author{Ricardo Martin-Brualla}
\affiliation{%
  \institution{Google Research}
}
\email{rmbrualla@google.com}

\author{Jason Lawrence}
\affiliation{%
  \institution{Google Research}
}
\email{jdlaw@google.com}

\author{Steven M. Seitz}
\affiliation{%
  \institution{University of Washington and Google Research}
}
\email{seitz@cs.washington.edu}

\begin{abstract}
    Many historical people were only ever captured by old, faded, black and white photos, that are distorted due to the limitations of early cameras and the passage of time. This paper simulates traveling back in time with a modern camera to rephotograph famous subjects. Unlike conventional image restoration filters which apply independent operations like denoising, colorization, and superresolution, we leverage the StyleGAN2 framework to project old photos into the space of modern high-resolution photos, achieving all of these effects in a unified framework. A unique challenge with this approach is retaining the identity and pose of the subject in the original photo, while discarding the many artifacts frequently seen in low-quality antique photos. Our comparisons to current state-of-the-art restoration filters show significant improvements and compelling results for a variety of important historical people.
Please go to \rurl{time-travel-rephotography.github.io} for many more results.

\end{abstract}



\begin{teaserfigure}
\hsize=\textwidth
\centering
\includegraphics[width=0.99\textwidth]{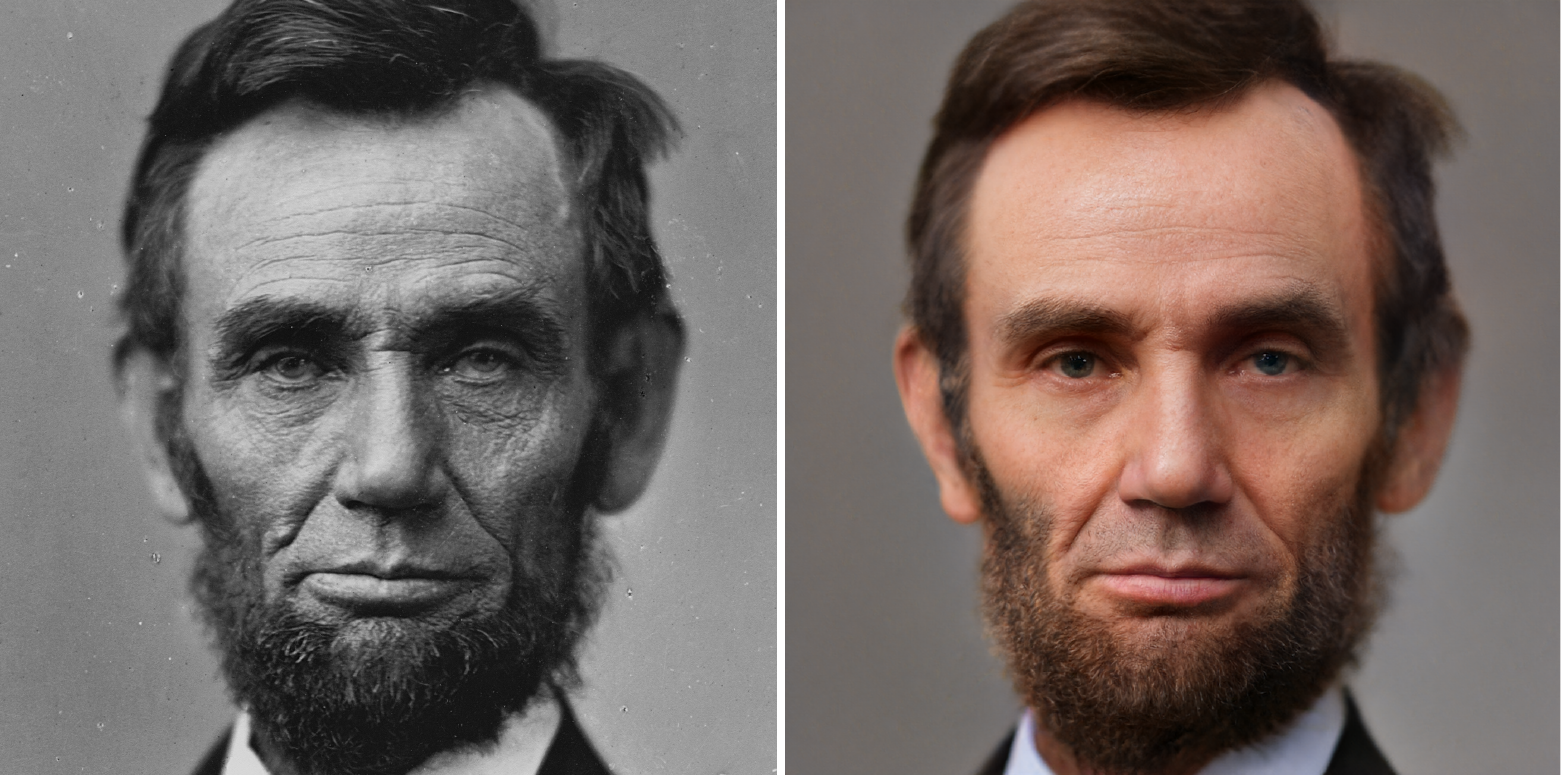}
\vspace{-1em}
\caption{
\small{Left: Antique film lacks red sensitivity, exaggerating wrinkles and darkening lips. Right: Our rendering of how Abraham Lincoln (c. 1863) would appear {\em rephotographed with a modern camera}. 
The input photo is from Mead Art Museum (public domain).
Images in all figures are best viewed digitally, and zoomed in to $1024\times1024$ to see details.
}
}
\label{fig:teaser}
\end{teaserfigure}

    \maketitle

\section{Introduction}

Abraham Lincoln’s face is iconic -- we recognize him instantly.  But what did he {\em really} look like?  Our understanding of his appearance is based on grainy, black and white photos from well over a century ago.  Antique photos provide a fascinating glimpse of the distant past. However, they also depict a faded, monochromatic world very different from what people at the time experienced. Old photos distort appearance in other less obvious ways.  For example, the film of Lincoln’s era was sensitive only to blue and UV light, causing %
cheeks to appear dark, and overly emphasizing wrinkles by filtering out skin subsurface scatter which occurs mostly in the red channel.  Hence, the deep lines and sharp creases that we associate with Lincoln’s face (Fig.~\ref{fig:teaser}) are likely exaggerated by the photographic process of the time.

To see what Lincoln really looked like, one could travel back in time to take a photo of him with a modern camera, and share that photo with the modern world.  Lacking a time machine, we instead seek to simulate the result, by projecting an old photo into the space of modern images, a process that we call {\em time-travel rephotography}. %

Specifically, we start with an antique photo as reference, and wish to generate the high resolution, high quality image that a modern camera would have produced of the same subject.  This problem is challenging, as antique photos have a wide range of defects due both to the aging process (fading and dust) and to limitations of early cameras, film, and development processes (low resolution, noise and grain, limited color sensitivity, development artifacts). A common approach is to try to restore the image, by applying a sequence of digital filters that attempt to undo these defects, e.g., noise removal, image deblurring, contrast adjustment, super-resolution, and colorization. A challenge with this approach is that the properties of old film and the aging process haven’t been fully characterized – hence, undoing them is an ill-posed problem.

\begin{figure*}[ht]
    \centering
    \includegraphics[width=0.9\linewidth]{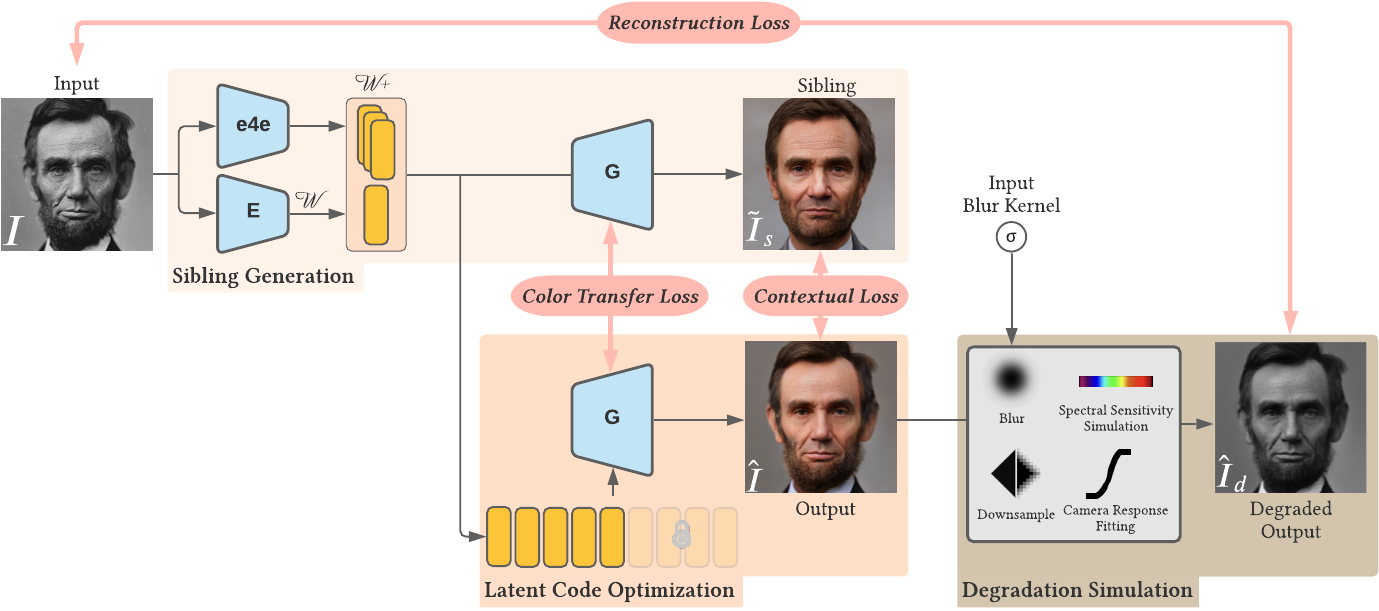}
    \vspace{-1em}
    \caption{Given an input antique photo, we generate a {\em sibling} in the \stylegantwo{} latent space by 
    \emph{style mixing}~\cite{karras2019style} the predictions from two feed-forward encoders, one that models face identity (\emph{e4e}~\cite{tov2021designing}) and another for face color ($E$).
    We then optimize the latent code of the sibling to match the input,
    after passing through a degradation model that simulates antique images, guided by the color, contrast, and skin textures of the sibling.
    }
    \label{fig:projection-diagram}
    \vspace{-1em}
\end{figure*}

Instead, we propose to project the antique photo into the space of modern images, using generative tools like 
\stylegantwo{}~\cite{karras2020analyzing}.
Unlike prior StyleGAN inversion methods~\cite{abdal2019image2stylegan, abdal2020image2stylegan++, richardson2020encoding, tov2021designing, zhu2020indomain, baylies2019}, our goal is not to reconstruct the input grayscale image, but to synthesize its missing colors, dynamic range and skin details while preserving the subject's identity. 
Our approach uses a physically-based \emph{film degradation} operator that simulates properties of antique cameras and the film aging process. This includes modeling the film's chromatic sensitivity, addressing for the first time different antique photographic emulsions (\emph{blue-sensitive}, \emph{orthochromatic} and \emph{panchromatic}~\cite{adams2018negative}). We also show that sharp photos with pleasing exposure and contrast can be achieved by explicitly modeling blur and non-linear camera response functions. Finally, to ensure natural color tones and high resolution face details, we synthesize a high-quality modern \emph{sibling} image in the \stylegantwo{} space to serve as an exemplar (Fig.~\ref{fig:projection-diagram}). We transfer the color and skin details from the sibling to our output using a contextual loss~\cite{mechrez2018contextual} and a novel \emph{color transfer loss} (Sec.~\ref{sec:method}).

We present results that show our approach, which addresses artifact removal, colorization, super-resolution, and contrast adjustment in one unified framework, consistently outperforms applying a sequence state-of-the-art image restoration filters. We also demonstrate compelling time-travel portraits of many well-known historical figures from the last two centuries, including presidents, authors, artists, and scientists (Fig.~\ref{fig:teaser},~\ref{fig:comparison-highres}). 
\xuan{Since we rely heavily on the face priors from \stylegantwo{},
our method is prone to the bias~\cite{salminen2020analyzing} of \stylegantwo{} and its training set. Such bias and our limitations are discussed in Sec.~\ref{sec:limitations}.}

\section{Related Work}
\label{sec:related-works}

Our method integrates concepts from the mature image restoration literature with modern learning-based approaches for modeling the space of high-resolution images of human faces.

\paragraph{Image Restoration.} Many prior methods address a \emph{single} type of image degradation, including denoising~\cite{buades2005non, elad2006image, dabov2007image, lefkimmiatis2017non, xie2012image, zhang2017beyond, zhang2017learning, zhang2018ffdnet, zhang2018residual}, deblurring~\cite{kupyn2018deblurgan, nah2017deep, xu2014deep, xu2014inverse, sun2015learning}, JPEG image deblocking~\cite{dong2015compression, guo2016building, wang2016d3}, and super-resolution~\cite{babacan2008total, dong2015image, kim2016accurate, ledig2017photo, menon2020pulse, tai2017memnet, yang2010image}. 
Some methods target specifically at faces, including deblurring~\cite{hacohen2013deblurring, pan2014deblurring} and super-resolution ~\cite{bulat2018super, grm2019face, menon2020pulse, ren2019face, shen2018deep}. 
To address restoring {\em multiple} artifacts, researchers have proposed using reinforcement learning~\cite{yu2018crafting} or attention-based mechanisms~\cite{suganuma2019attention} to select the best combination of restoration operations to apply to each image.
The work by Wan et al.~\shortcite{wan2020} also restores portraits suffering from multiple artifacts. However, it does not do super resolution and its restoration quality degrades at high resolution of $1024\times 1024$ as evaluated in Sec.~\ref{sec:experiments}.
\xuan{The concurrent work by Wang et al.~\shortcite{wang2021gfpgan} also leverages generative facial prior. They propose to incorporate such prior into joint face restoration and color enhancement using spatial feature transform layers.
Nevertheless, none of the aforementioned techniques address colorization.}

Colorization research can be categorized into scribble-based, 
exemplar-based and learning-based methods. Early work~\cite{huang2005adaptive, levin2004colorization, luan2007natural, qu2006manga, sykora2009lazybrush, yatziv2006fast} 
manually specify target colors for parts of the image via sparse {\em scribbles} drawn on the image.
To reduce user involvement, an alternative is to transfer color statistics from a (manually-specified) reference image
\cite{charpiat2008automatic, chia2011semantic, gupta2012image, he2018deep, ironi2005colorization, liu2008intrinsic, welsh2002transferring}.
\xuan{Identifying a suitable reference image, however, is a research topic by itself. Automatic methods often involve a complicated image retrieval system~\cite{chia2011semantic}. We simplify this process by predicting a high-quality \stylegantwo{} sibling as the reference.}
Most related are {\em fully automated} colorization methods that use machine learning on
a large dataset~\cite{cheng2015deep, deshpande2015learning, iizuka2016let, isola2017image, larsson2016learning, zhang2016colorful, zhang2017real, zhao2018pixel}.
We compare with many of these methods in Sec.~\ref{sec:experiments}.

Despite the rapid progress in these individual areas, no prior works have addressed restoration, colorization and super-resolution in a single framework. 
We demonstrate that doing so produces better results than applying a sequence of state-of-the-art techniques.

\begin{figure*}[ht]
    \centering
    \setlength{\tabcolsep}{0pt} 
    \def\imW{0.167\linewidth}
    \begin{tabular}{p{\imW}p{\imW}p{\imW}p{\imW}p{\imW}p{\imW}}
        \multicolumn{6}{c}{\includegraphics[width=\linewidth]{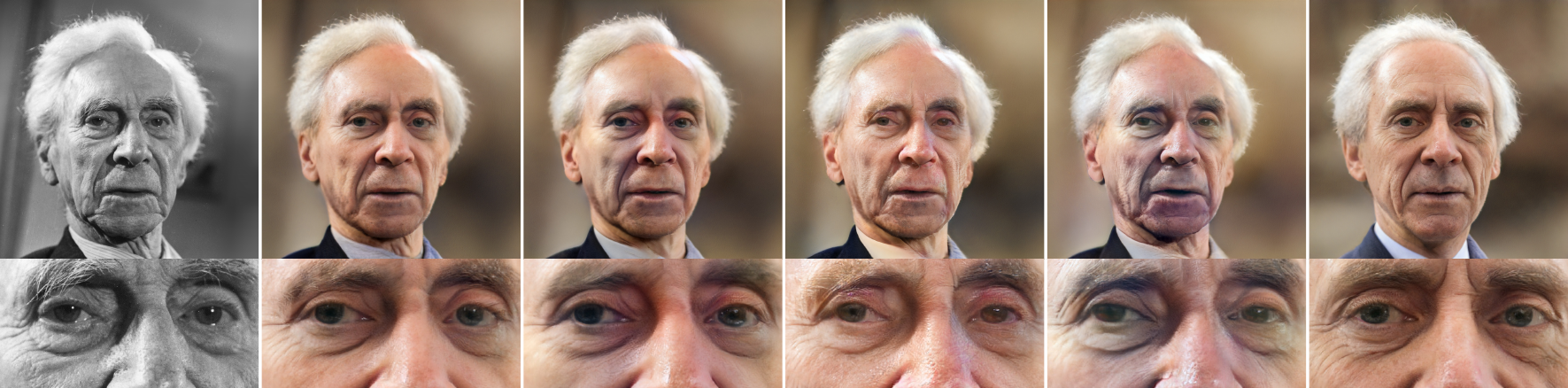}}\\
        \multicolumn{1}{c}{(a) Input} &  \multicolumn{1}{c}{(b) Ours} & \multicolumn{1}{c}{(c) w/o $\mathcal{L}_{color}$} & \multicolumn{1}{c}{(d) w/o $\mathcal{L}_{ctx}$} &  \multicolumn{1}{c}{(e) w/o $\mathcal{L}_{ctx}+\mathcal{L}_{color}$} & \multicolumn{1}{c}{(f) Sibling} \vspace{-2.5em} \\
        & & & & &
    \end{tabular}
    \caption{Impact of each loss for transferring  color and details from the sibling.
    \xuan{
    In all these examples, the sibling code is used as an initialization and its fine codes (128 - 1024) are kept unchanged during optimization. Yet, low-frequency color artifacts appear without $\mathcal{L}_{color}$. Skin 
    and eye details are poorly reconstructed without $\mathcal{L}_{ctx}$.   %
    }
    Input image: Bertrand Russell (1872 - 1970) from BBC Photo Library.
    }
    \label{fig:loss-ablation}
    \vspace{-1.3em}
\end{figure*}

\paragraph{Face embedding.}
Karras et al. \shortcite{karras2019style, karras2020analyzing} introduced
the StyleGAN framework for synthesizing high resolution human faces from a latent space.
Projecting or {\em embedding} a face image to the latent space is an active research topic.
Current methods fall into three categories: 1) optimizing latent vectors to best fit the input image~\cite{abdal2019image2stylegan, abdal2020image2stylegan++}, 2) training an \emph{encoder network} that estimates a latent coordinate from an input image~\cite{richardson2020encoding, tov2021designing}, and 3) hybrid methods~\cite{zhu2020indomain, baylies2019}
that use an encoder network to initialize an optimization, similar to our approach.
In contrast to our work, all of these prior methods seek to reconstruct the input directly and do not address the restoration problem, i.e., they would seek embeddings that preserve an antique image's monochrome, blurry, and low-contrast properties.
Directly extending these prior methods to overcome this limitation by inserting an image degradation operator into the loss function results in a poorly conditioned optimization problem that can quickly converge to a bad result (Fig.~\ref{fig:loss-ablation}e), even with a good starting point \cite{zhu2020indomain, baylies2019}.
Our use of a sibling image
is critical for obtaining a well conditioned optimization problem that reliably converges to a high-quality result.
We note that Menon et al.~\shortcite{PULSE_CVPR_2020} addressed the specific problem of upsampling. However, they focus on high-ratio upsampling where the identity of input images are hardly recognizable. Similarly, Yang et al.~\shortcite{Yang2021GPEN} uses GAN priors to blindly restore images with extreme low resolutions without explicitly constraining the identity. Recently, Pan et al.~\shortcite{pan2020dgp} exploits deep generative prior for independent image colorization and super resolution tasks, where they further finetune the generator together with the latent code to reduce the gap between 
the training and testing images. However, none of these works take into account the complicated real-world degradations of antique photos such as the chromatic sensitivity and contrast adjustment that are critical for high-quality antique photo restoration.

\section{Problem Statement}
\label{sec:problem-formulation}

Our goal is to simulate traveling back in time and rephotographing historical figures with a modern camera.
We call this {\em time-travel rephotography}, 
adapting the term 
{\em rephotography} which traditionally means ``the act of repeat photography of the same site, with a time lag between the two images'' \cite{rephotoWiki}. %
We mainly focus on portraits recorded a century ago, shortly after cameras were invented in the late 1800s, which
are challenging to restore due to loss of quality through the aging process and limitations of early film, but our method operate on more recent photos as well.

Photographic film has evolved significantly since its invention.
The first light-sensitive emulsions were sensitive only to blue and ultraviolet light~\cite{Newhall1982history}. \emph{Orthochromatic} emulsions~\cite{Newhall1982history}, introduced in 1873, provided
sensitivity to both green and blue light. 
Photographic portraits from these eras rendered skin poorly, artificially darkening lips and exaggerating wrinkles, creases, and freckles, due to the lack of red sensitivity. In particular, they underestimate the effect of subsurface scattering, which gives skin its characteristic smooth appearance~\cite{jensen}.
{\em Panchromatic} film, sensitive to red, green, and blue first appeared in 1906, yet orthochromatic films remained popular through the first half of the 20th century \cite{filmWiki}.

To simulate rephotographing historical people with a {\em modern camera}, we must account for these differences in color sensitivity of antique film, in addition to blur, fading (poor exposure, contrast), noises, low resolution, and other artifacts of antique photos.

\section{Method}
\label{sec:method}

We seek to synthesize a {\em modern photo} of a historical person, using an antique black-and-white photo as reference.  
Our approach is based on the idea of {\em projecting} the antique photo into the space of modern high-resolution color images represented by the StyleGAN2 generative model~\cite{Karras2019stylegan2}.

Similar to previous techniques~\cite{baylies2019}, we optimize the latent space of StyleGAN2 to synthesize an image. 
However, directly fitting the antique image would reproduce a grainy black and white result, we instead reconstruct an image of the person without any artifact caused by antique negatives or the film aging process.

A first step is to convert the StyleGAN2 output to grayscale before comparing with the antique input image. 
This naive approach is poorly constrained, since multiple colors can correspond to the same grayscale output, and thus leads to unrealistic colorized results (Fig.~\ref{fig:loss-ablation}e). 
Therefore, we employ an additional exemplar image as a reference,
that has similar facial features as the input, yet contains high frequency details and natural color and lighting.
Sec.~\ref{sec:sibling_encoder} explains how we compute such an exemplar automatically.
We call it the \emph{sibling} image, as it resembles characteristics of the input while having a different identity. Sec.~\ref{sec:sibling_color_transfer} introduces losses used in optimization that constrain our rephotographed output to retain the contrast, color and high-frequency details present in the sibling.

To further reduce the perceived identity gap between the input image and the modern portrait, we design reconstruction losses suited for antique images (Sec.~\ref{sec:reconstruction-losses-for-antique-images}).
A key contribution is the proposed degradation module, that simulates the image formation model in antique photos, and is applied to the StyleGAN2 result before comparing it with the input antique photo.
The degradation module accounts for different types of film substrate, camera response curves, image blur, and low resolution, which altogether leads to improved rephotography results. We provide details on the latent code optimization in Sec.~\ref{sec:latent-code-optimization} and a system overview in Figure~\ref{fig:projection-diagram}.

\subsection{Sibling Encoders}
\label{sec:sibling_encoder}
Given a low-resolution grayscale reference image as input $I$, we seek
to generate a high-resolution color {\em sibling} image $\tilde{I}_s$ that has photo-realistic colors and preserves facial features in the original input.
The state-of-the-art StyleGAN inversion method \emph{e4e}~\cite{tov2021designing} can embed an input image into the $\mathcal{W+}$ space~\cite{abdal2019image2stylegan} (18 different 512-dimensional \stylegantwo{}  $\mathcal{W}$ codes) with reasonably well-preserved identity. However, when applied to antique photos, such inversion methods will also preserve the artifacts, such as blur and lack of color. An alternative approach is to train a feed-forward encoder $E$, aimed specifically at predicting a high-quality (i.e. realistic skin color and details) \stylegantwo{} embedding from an antique image. Avoiding the transfer of artifacts is possible by limiting the expressiveness of the embedding, i.e. predicting an embedding in $\mathcal{W}$ instead of $\mathcal{W+}$. As a result, this embedding is also worse at preserving the subject's identity.

Our method combines the best of both worlds by \emph{style mixing}~\cite{karras2019style} the predictions from \emph{e4e} and from $E$. Specifically, we use \emph{e4e}'s $\mathcal{W+}$ prediction for the first 10 coarse style codes and duplicate our $\mathcal{W}$ prediction for the other 8 fine codes.
The mixed $\mathcal{W+}$ codes are then converted to the sibling image using the pre-trained \stylegantwo{} generator~\cite{Karras2019stylegan2}.
Our encoder $E$ is trained using random samples of \stylegantwo{} $\mathcal{W}$ latent codes and their corresponding images, downsampled  to $256 {\times} 256$ and converted to grayscale based on the emulsion type (see details in Sec.~\ref{sec:reconstruction-losses-for-antique-images}).

\subsection{Sibling Color And Detail Transfer}
\label{sec:sibling_color_transfer}

To further constrain colors and skin details to match the sibling (Fig.~\ref{fig:loss-ablation}),
we introduce a \textbf{\emph{color transfer loss}} $\mathcal{L}_{color}$ that enforces the distribution of the output colors from StyleGAN2's \texttt{ToRGB} layers to be similar to those of the sibling image. We use a formulation inspired by style loss~\cite{gatys2016image}, and apply to \texttt{ToRGB} layers outputs.
In our implementation, we use the mean-squared distance between elements in the covariance matrices.

Although $\mathcal{L}_{color}$ encourages matching the overall style of the image, details like skin texture were still not transferred properly (Fig.~\ref{fig:loss-ablation}d). One could have a reconstruction loss between the sibling and the output to encourage detail synthesis, 
but such a loss would be very sensitive to misalignment between the sibling and the \stylegantwo{} result, and encourage identity shift.
We thus introduce a \textbf{\emph{contextual loss}} $\mathcal{L}_{ctx}$~\cite{mechrez2018contextual} between the VGG features of $\tilde{I}_s$ and $\hat{I}$. %
$\mathcal{L}_{ctx}$ compares each feature in $\hat{I}$ to \emph{all} features in $\tilde{I}_s$, and minimizes the distance to only the most similar one. This allows for the transfer of high-frequency details without requiring precise alignment.

\subsection{Reconstruction Losses for Antique Images}
\label{sec:reconstruction-losses-for-antique-images}

Rather than fitting the antique input image exactly, we seek a loss that helps preserve the input's identity while being robust to defects of antique photos.
We approach this by introducing a \textbf{\emph{reconstruction loss}} $\mathcal{L}_{recon}$ that applies a series of modifications to the StyleGAN2 result before comparing it to the antique input image.
We define $\mathcal{L}_{recon}$ as a loss between the input image $I$ and a modified version of \stylegantwo{} output $\hat{I}_d = \mathcal{D}(\hat{I})$, where $\mathcal{D}$ is a degradation process, which attempts to make the generated image appear as if it were captured by an antique camera. In the following, we design $\mathcal{D}$ to account for the spectral sensitivity of early negatives, different camera response functions, possible image blur, and resolution differences.

\begin{figure}
    \centering
    \def\imW{0.33\linewidth}
    \setlength{\tabcolsep}{1pt}
    \begin{tabular}{ccc}
        \includegraphics[width=\imW]{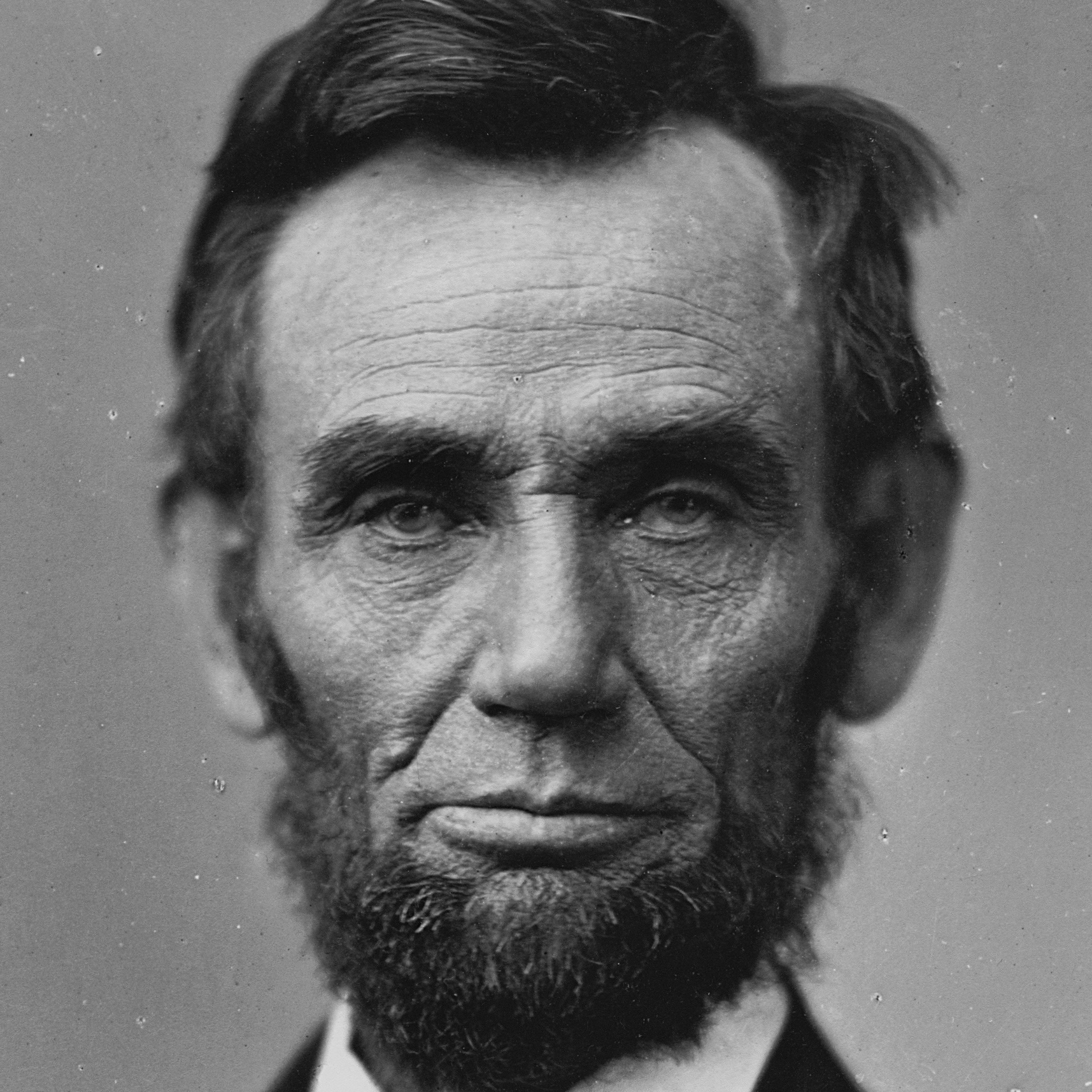}&\includegraphics[width=\imW]{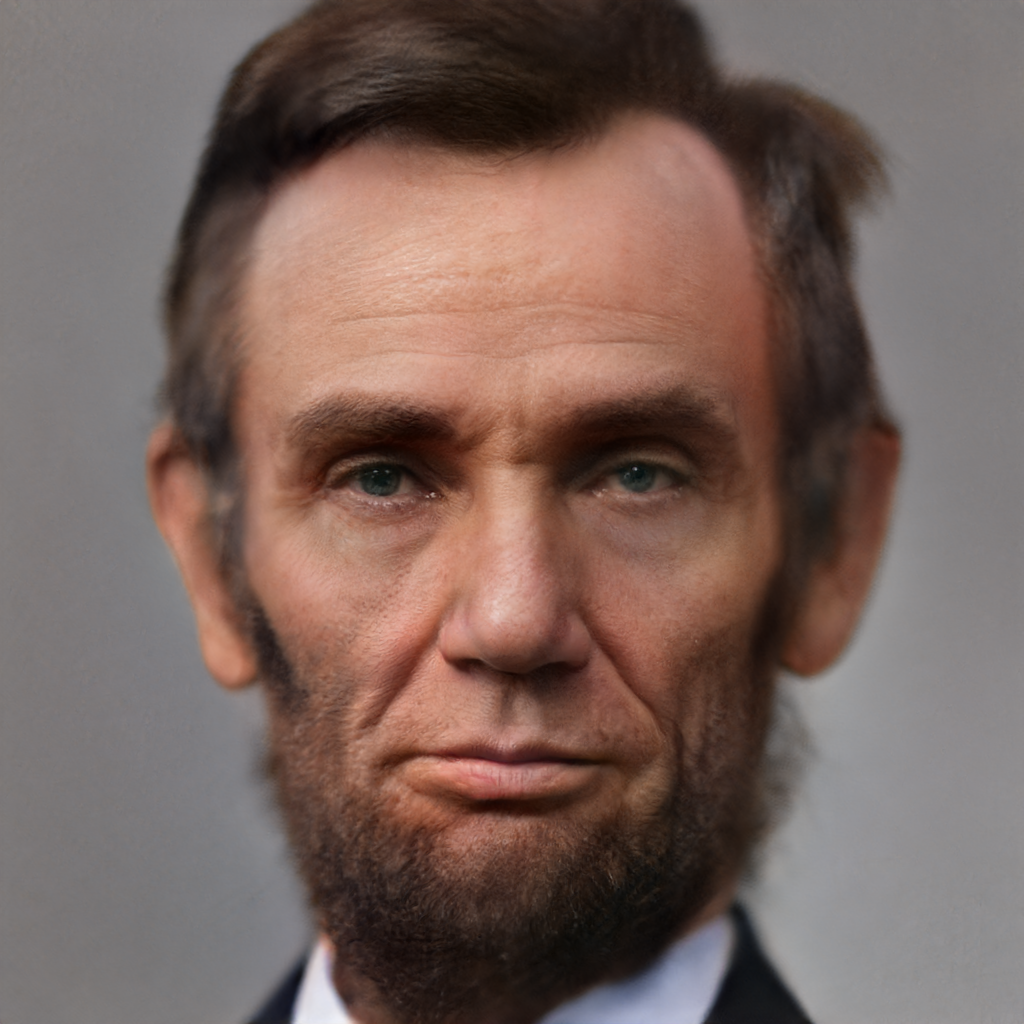}&\includegraphics[width=\imW]{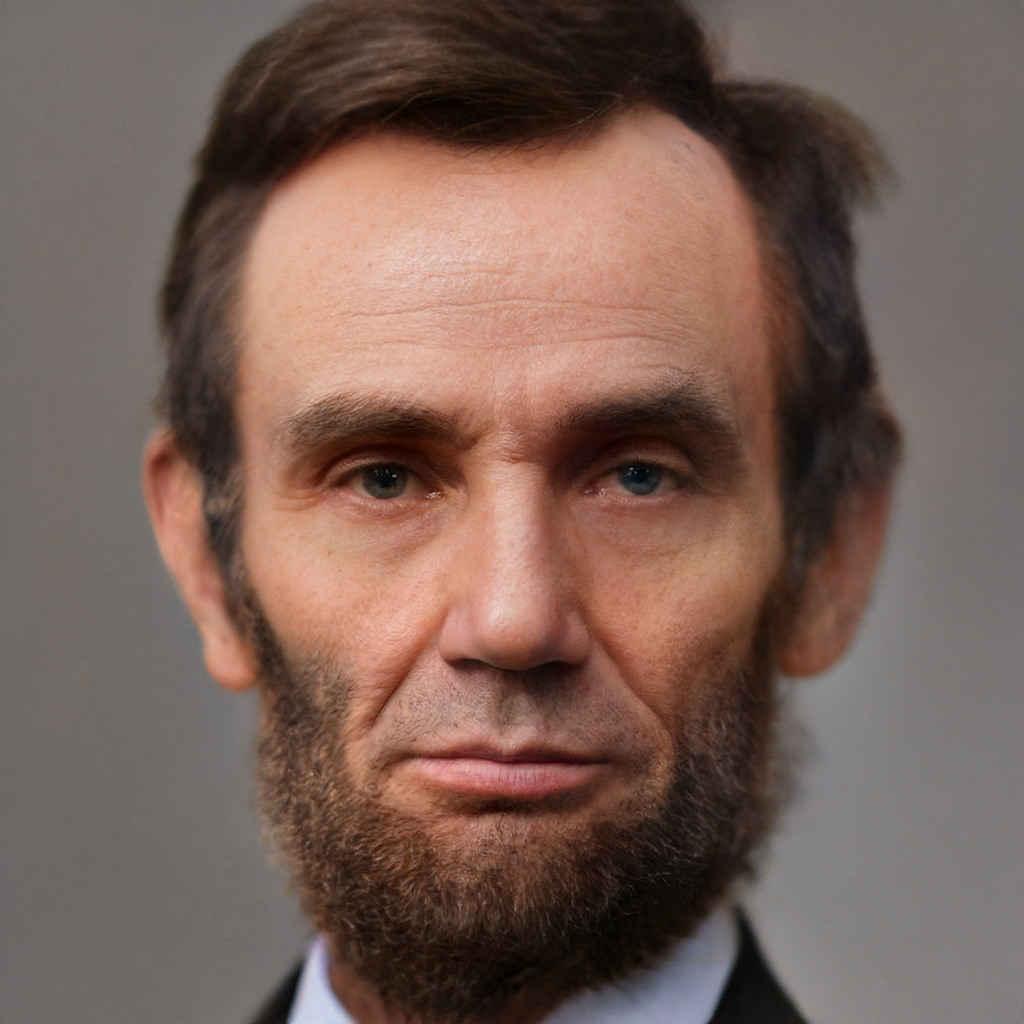} \vspace{-4pt} \\
        \small{Input}& \small{Panchromatic} & \small{Blue-sensitive} \\
    \end{tabular}
    \vspace{-1.5em}
    \caption{
        Abraham Lincoln c. 1863, when the negatives were sensitive only to blue light. The panchromatic reconstruction has exaggerated wrinkles and unnatural colors. Input image: Mead Art Museum (public domain). %
    }
    \vspace{-1em}
    \label{fig:spectral-sensitivity}
\end{figure}

\begin{figure}
    \centering
    \def\imW{0.248\linewidth}
    \setlength{\tabcolsep}{0.5pt}
    \begin{tabular}{cccc}
         \includegraphics[width=\imW]{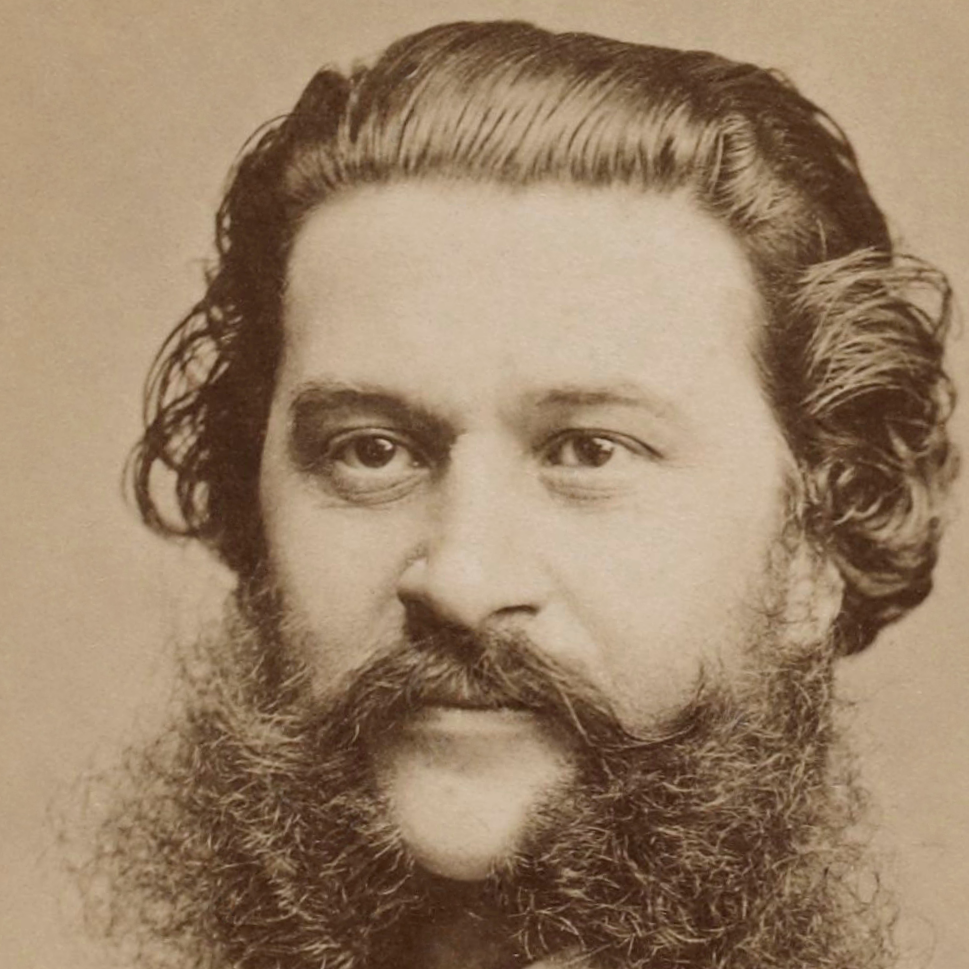} & \includegraphics[width=\imW]{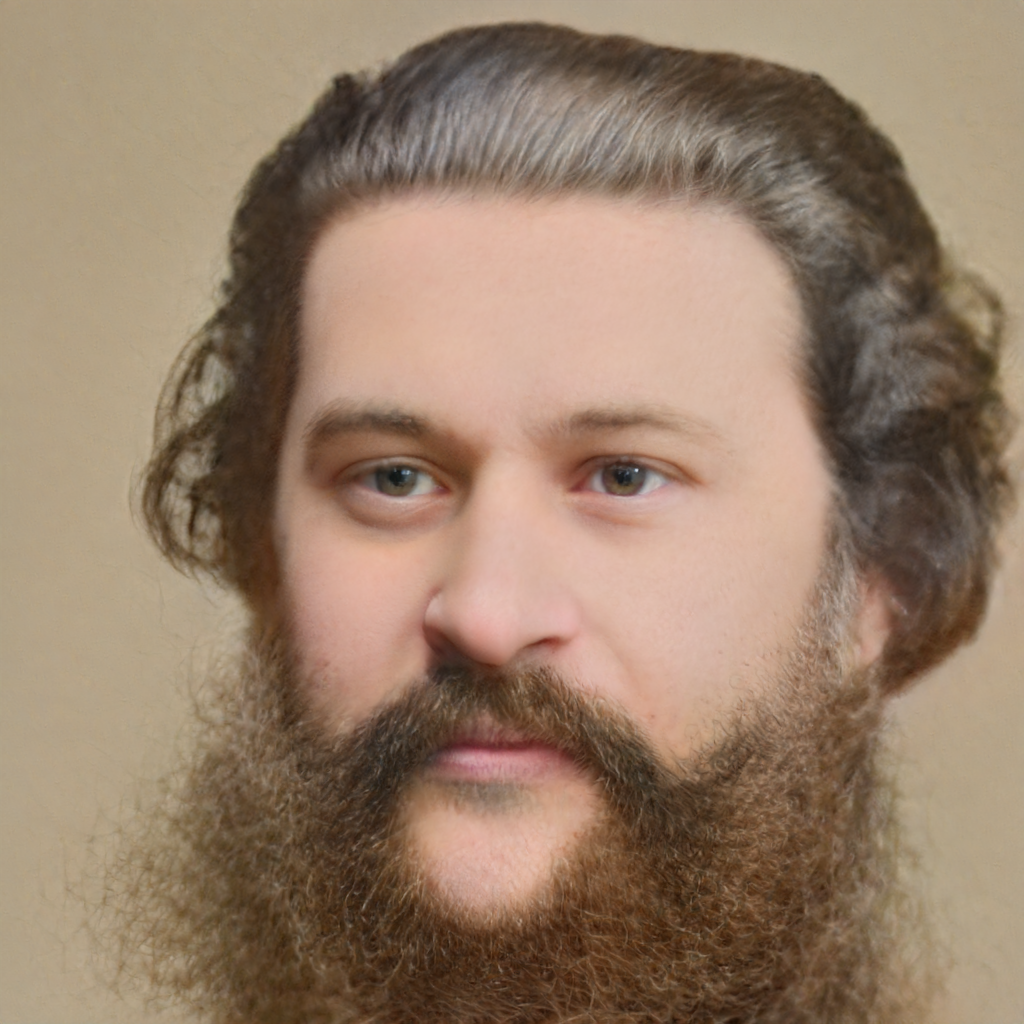} & \includegraphics[width=\imW]{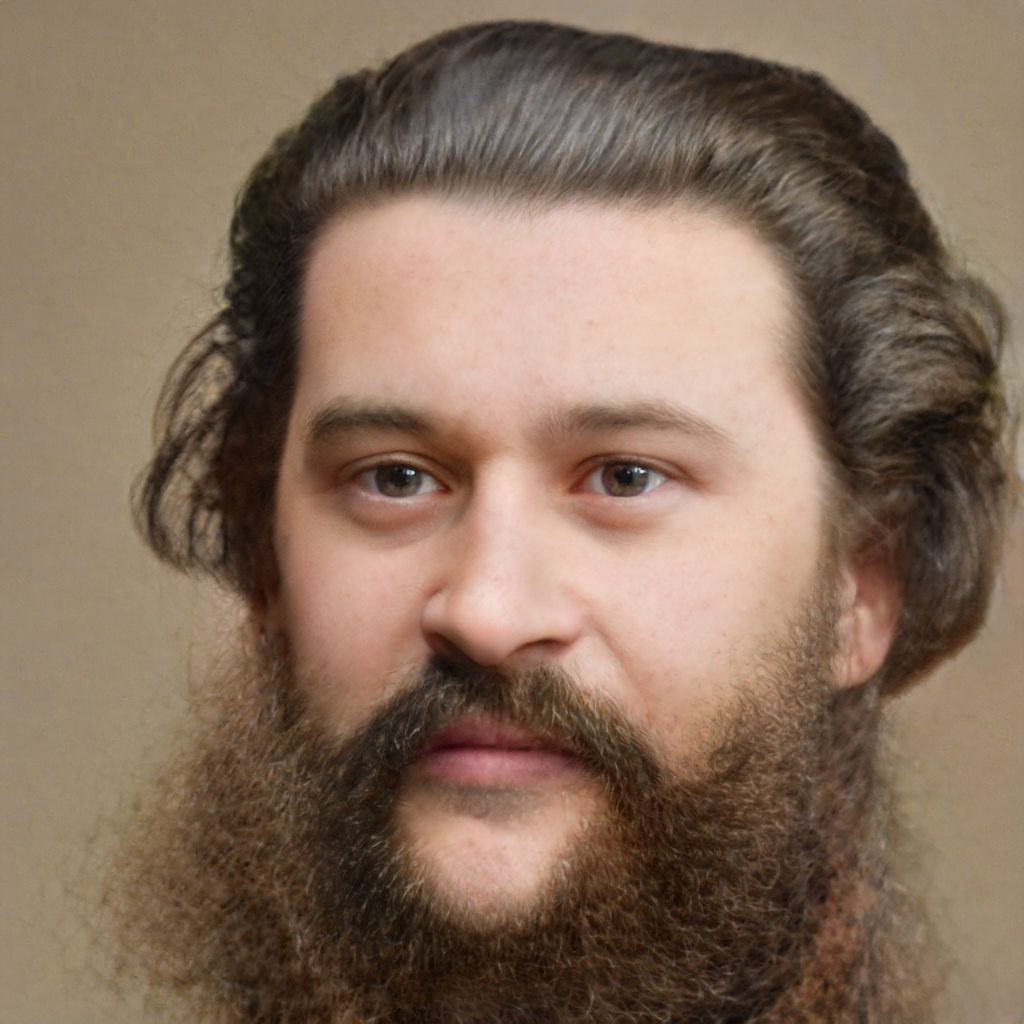} & \includegraphics[width=\imW]{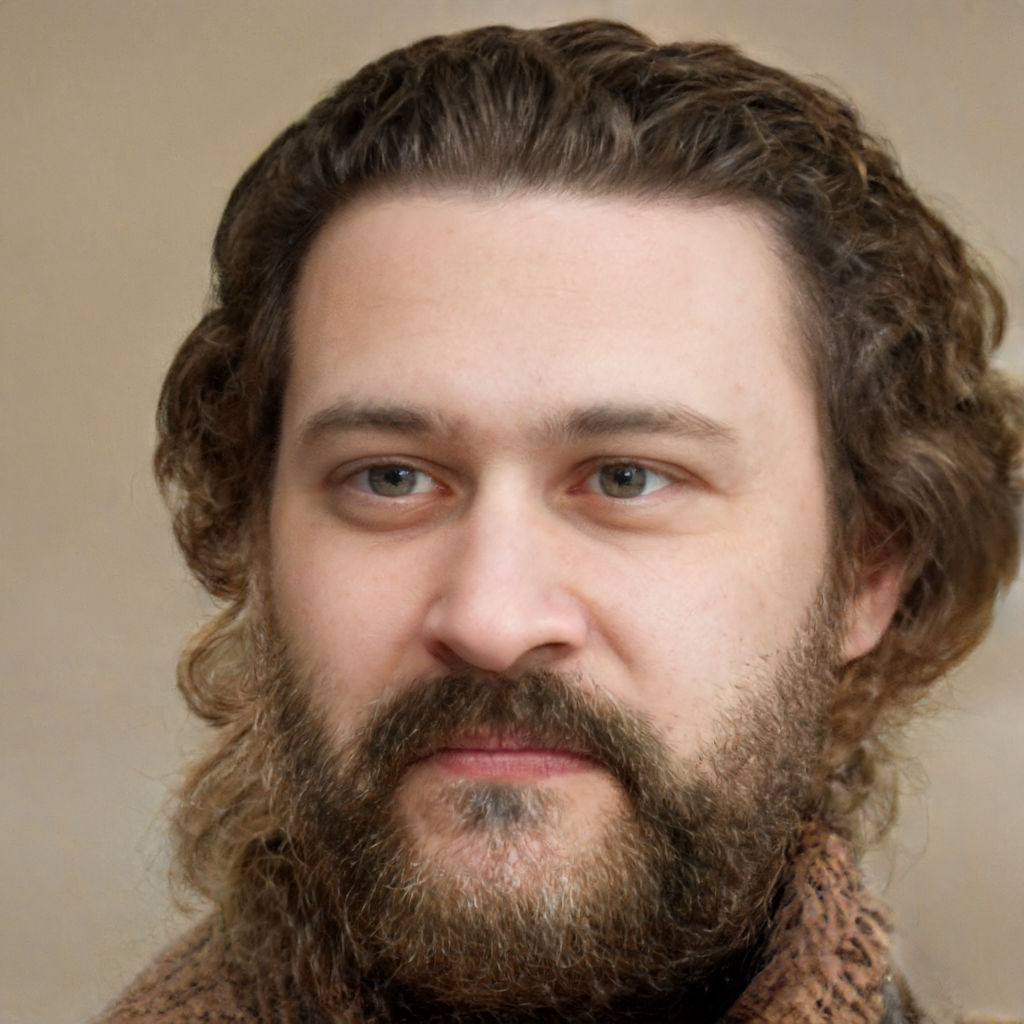} \vspace{-1mm}\\
         Input & w/o CRF-F & with CRF-F & Sibling \vspace{-1pt}
    \end{tabular}
    \vspace{-1.2em}
    \caption{Camera response function fitting (CRF-F) helps low-contrast images recover a wider dynamic range by making their contrast and exposure close to the sibling. Input image: Johann Strauss II (1899) from gallica.bnf.fr / BnF. %
    }
    \label{fig:camera-response}
    \vspace{-1.5em}
\end{figure}
\mypar{Antique Film Spectral Sensitivity}
We convert \stylegantwo{}'s output $\hat{I}$ to grayscale, denoted as $\hat{I}_g=\mathcal{G}(\hat{I})$ where $\mathcal{G}$ is the grayscale conversion process.
The grayscale conversion must accommodate for the unique sensitivity of early film which is far more sensitive to blue light than red.
In particular, we extract the blue channel for \emph{blue-sensitive} photos, average 
the blue and green intensities ($0.5\cdot (G + B)$) to approximate \emph{orthochromatic} photos~\cite{geigel1997model}, and use standard grayscale conversion ($0.299\cdot R+0.587\cdot G+0.114\cdot B$) for \emph{panchromatic} photos. As shown in Fig.~\ref{fig:spectral-sensitivity}, choosing the right film 
model affects the result quality. %
When multiple spectral sensitivity types are possible (e.g. based on the photo's capturing time), 
the user can choose the one that produces the best result.
\vspace{2pt}

\mypar{Camera Response Function}
We model the unknown Camera Response Function (CRF) of the input old photo as 
    $a + b \hat{I}_g ^\gamma$
where $a, b$ and  $\gamma$ are the bias, gain and gamma parameters to be optimized. During optimization, we initialize with $a=0$ and $b=\gamma=1$.
To improve convergence, we pre-align the appearance of the input to be closer to the sibling using \emph{histogram matching}~\cite{burger2016digital}, i.e., 
we convert the sibling image to grayscale $\mathcal{G}(\tilde{I}_s)$, apply
histogram transform to the input image to match the grayscale sibling (in the face region only) and produce $I'$. We then set $I'$ instead of $I$ to be the reconstruction target $\mathcal{L}_{recon}(I', \hat{I}_d)$. 
See the supplement for more details. 
We observe in Fig.~\ref{fig:camera-response} that when the input images suffer from poor exposure and contrast, CRF fitting helps avoid transferring these artifacts to the output image by bringing the exposure and contrast closer to those of the sibling. 
\vspace{2pt}

\begin{figure}[t]
    \centering
    \setlength{\tabcolsep}{1pt}
    \def\imW{0.33\linewidth}
    \def\imH{0.26\linewidth}
    \begin{tabular}{p{\imW}p{\imW}p{\imW}}
        \multicolumn{3}{c}{\includegraphics[width=\linewidth]{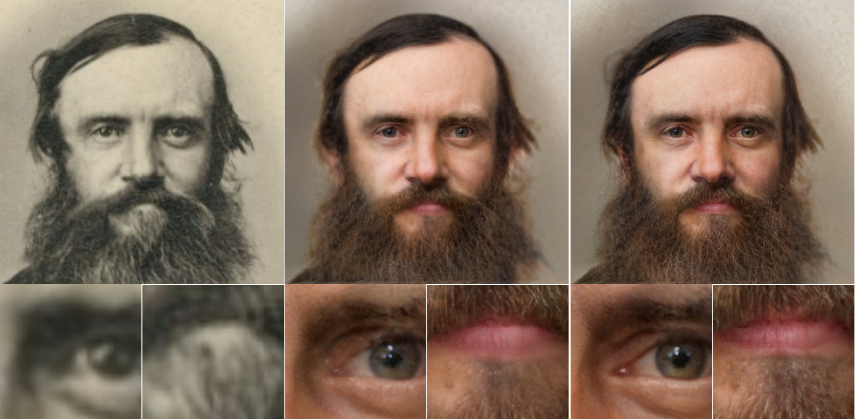}}
        \vspace{-1mm}\\
        \multicolumn{1}{c}{\small Input \quad} &  \multicolumn{1}{c}{\small w/o blur simulation} & \multicolumn{1}{c}{\small w/ blur simulation} \vspace{1mm} \\
        & & 
    \end{tabular}
    \vspace{-3em}
    \caption{Impact of blur simulation in our reconstruction loss. 
    Best viewed full screen. 
    Input image: John McDouall Stuart (c. 1860) from the State Library of South Australia~\shortcite{slsa}: B 501.
    }
    \label{fig:blur-simulation}
    \vspace{-2em}
    \vspace{2mm}
\end{figure}

\mypar{Blur}
We finally apply a Gaussian blur with a user-provided standard deviation $\sigma$ to obtain the final degraded result $\hat{I}_d$. Values between 0 and 7 work well in our experiments.
This blur accounts for the loss of details from the aging, scanning, and capture process (e.g., defocus, low film quality, etc.).
Fig.~\ref{fig:blur-simulation} illustrates the benefit of simulating blur during optimization.

\vspace{2pt}
\mypar{Reconstruction Loss}
Using the degradation process $\mathcal{D}$ outlined above, we now define our reconstruction losses. 
To capture the face identity, we \emph{downsample} both $I'$ and $\hat{I}_d=\mathcal{D}(\hat{I})$ from $1024{\times}1024$ to $256{\times}256$ and compute a perceptual loss between these downsampled images using a combination of VGG~\cite{simonyan2014very} and VGG-Face~\cite{Parkhi15} features. We add additional constraints to the eye region, which plays an essential part in human perception of faces. Specifically, we \emph{downsample} $\hat{I}_d$ to the original input resolution of $I'$, crop it and $I'$ in the eye regions to get $\hat{I}_d^{eye}$ and $I^{eye}$. We apply VGG-based perceptual loss to reconstruct these eye crops.
The complete reconstruction loss is:
\vspace{-5pt}
\begin{equation}
   \begin{split}
    \mathcal{L}_{recon} = & ~\lambda_{\mathrm{vgg}} \mathcal{L}_{\mathrm{vgg}}(f(I'), f(\hat{I}_d)) +\lambda_{\mathrm{face}} \mathcal{L}_{\mathrm{face}}(f(I'), f(\hat{I}_d)) \\
     & ~+\lambda_{eye} \mathcal{L}_{\mathrm{vgg}} (I^{eye}, \hat{I}_d^{eye}), 
\end{split}
\end{equation}
where $f(\cdot)$ is the $4\times$ downsampling operator to $256{\times}256$.

\subsection{Latent Code Optimization}
\label{sec:latent-code-optimization}
Similar to previous methods~\cite{baylies2019, zhu2016generative},  %
we first initialize with an encoder output (in our case, the sibling's latent codes), and then optimize the latent codes, the CRF parameters, and the \stylegantwo{} per-layer noise maps. These are optimized to minimize the following loss:
\begin{equation}\mathcal{L}_{recon} + \lambda_{color}\mathcal{L}_{color} + \lambda_{ctx}\mathcal{L}_{ctx} + \lambda_{noise}\mathcal{L}_{noise}\end{equation} where $\mathcal{L}_{noise}$ is a noise map regularization loss~\cite{Karras2019stylegan2}.
In StyleGAN2~\cite{Karras2019stylegan2}, the latent codes in $\mathcal{W+}$ are used at different scales from $4{\times}4$ to $1024{\times}1024$. These codes roughly correspond to different perceptual aspects of an image. The coarser spatial codes 
determine the overall structure of the face (identity, pose, expression %
etc.), whereas the finer layers 
encode aspects like skin tone, skin texture, and lighting. 
We leverage the expressiveness of $\mathcal{W+}$ space and optimize codes for layers up to $64 {\times} 64$, which are sufficient to capture the identity and facial features. The finer spatial codes are copied from those of the sibling.

\subsection{Implementation Details}
\label{sec:implementation}
Our method, including sibling computation and latent code optimization, takes about $10$ minutes on one NVIDIA TITAN Xp GPU to produce a $1024 {\times} 1024$ result. 

\mypar{Latent Code Optimization}
Rather than optimizing latent code layers (4-64) simultaneously, we achieve better results by first optimizing the coarse codes (4-32) for 250 iterations to obtain an intermediate result $\hat{I}_{32}$. Then, we set $\hat{I}_{32}$ as our new sibling to be used in the color transfer and contextual losses, and jointly optimize latent codes of resolution 4 to 64 for another 750 iterations, producing the final output $\hat{I}$. 
Note that the color transfer loss is only enforced on the \texttt{ToRGB} layers of the latent codes being optimized.
To navigate the latent space more comprehensively, we also add ramped-down noises to the latent codes as in the projection method of StyleGAN2~\cite{Karras2019stylegan2}.
We use the RAdam optimizer~\cite{liu2019radam} with default parameters and learning rates $0.1$ for the style codes and $0.01$ for the camera response function parameters. The weights of each loss are $\lambda_{vgg}{=}1$, $\lambda_{face}{=}0.3$, $\lambda_{ctx}{=}\lambda_{eye}{=}0.1$, $\lambda_{color}{=}10^{10}$, $\lambda_{noise}{=}5\times10^9$. 
See the supplement for details on the specific layers used for all the losses.

\mypar{Sibling Encoder ($\mathbf{E}$)} For each film type described in Section~\ref{sec:reconstruction-losses-for-antique-images}, we train a sibling ResNet18 encoder $E$~\cite{he2016deep} using $16,128$ StyleGAN2-generated samples that are converted to grayscale accordingly.
We use an L1 loss between the predicted and ground truth latent codes. We apply color jitter, contrast and exposure augmentations during training. More details are in the supplement.

\section{Experiments}
\label{sec:experiments}
\begin{figure*}
    \centering
    \resizebox{\linewidth}{!}{
    \begin{tabularx}{\linewidth}{YYYY}
        \multicolumn{4}{c}{\includegraphics[width=0.98\linewidth]{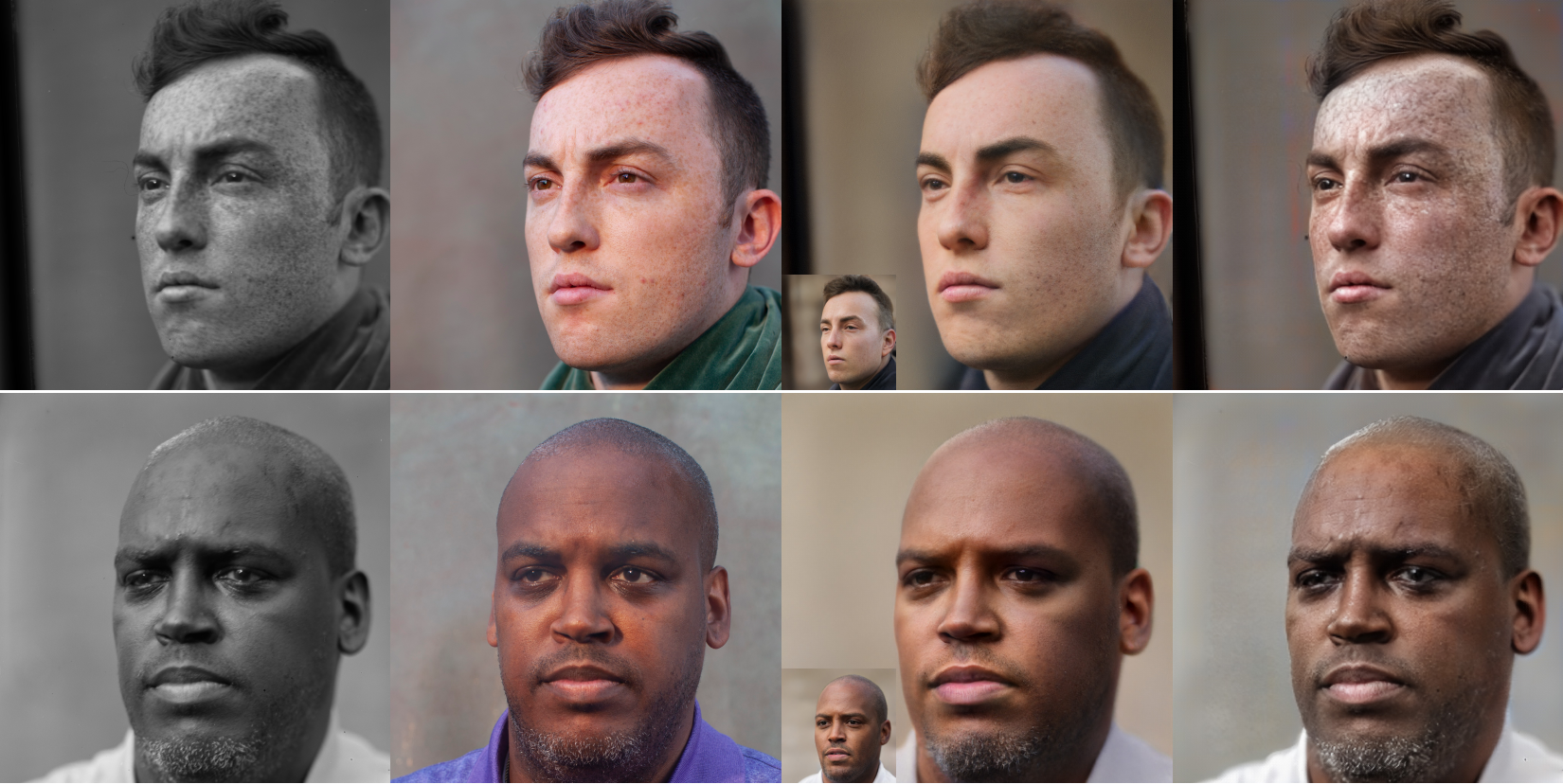}}\\
        (a) Input (Dry Plate) & (b) RGB (DSLR) &  (c) Ours & (d) Zhang (FFHQ)
    \end{tabularx}
    }
    \vspace{-1em}
    
    \caption{Comparisons of restored images from an antique photographic process with the ground truth color photos. 
    The same subject is captured 
    under the same lighting with both (b) a modern DSLR camera and an antique photographic process -- (a) \emph{gelatin dry plate}~\cite{wiki-dry-plate} \copyright Nick Brandreth, which is sensitive to both blue and UV light. We compare results from (c) our approach against the strongest baseline pipeline (d) \emph{Zhang (FFHQ)}. Our results tone down the exaggerated pimples, freckles, and skin specularity due to blue sensitivity and produces a more natural skin texture.
    }
    \label{fig:old_vs_new}
    \vspace{-1.5em}
\end{figure*}

\paragraph{Dataset.} We evaluate our methods on three sets of images. The first consists of a hand-picked set of photos that showcase the most interesting portraits of historical figures. We use this set for visualization purposes. For our second image set, to provide a fair and comprehensive evaluation, we propose a testing benchmark called the \emph{Historical Wiki Face Dataset}. This dataset is collected in an objective manner by automatically crawling Wikipedia and removing unsuitable samples. %
This set is very diverse, covering various styles and ethnic groups of important historical people. We use this benchmark for testing purposes, i.e., the user study.
\xuan{In our third set, we aim to compare the restored images with ground truth color photos. We asked a professional photographer
\emph{Nick Brandreth}~\shortcite{nick-brandreth} 
to reproduce a popular antique photographic process, the \emph{gelatin dry plate}~\cite{wiki-dry-plate}, which is sensitive to bothahblue and UV light. He captured the same subjects under the same outdoor lighting in black and white with the dry plate and in color with the modern DSLR camera from similar viewpoints (Fig.~\ref{fig:old_vs_new}).}
Please see our supplement for results from all three image sets.

\begin{figure*}[th]
    \centering
    \setlength{\tabcolsep}{3pt}
    \def\imW{0.155\linewidth}
    \begin{tabular}{cccccc}
        \includegraphics[width=\imW]{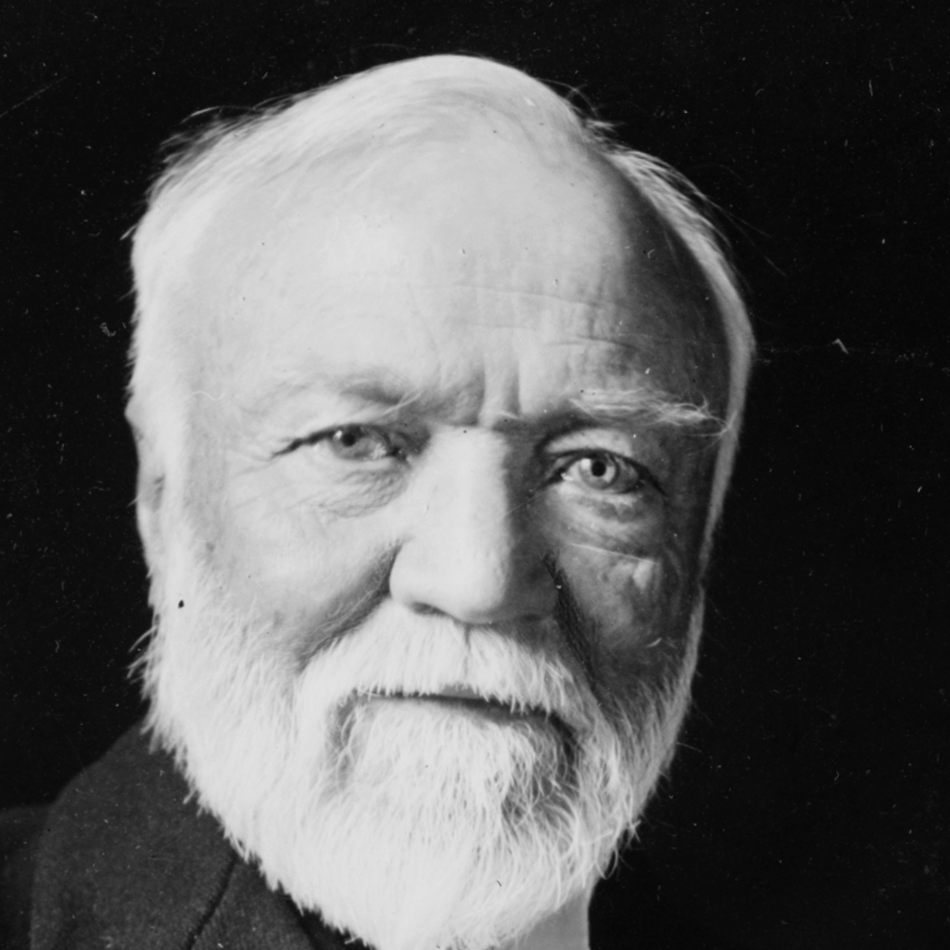} & \includegraphics[width=\imW]{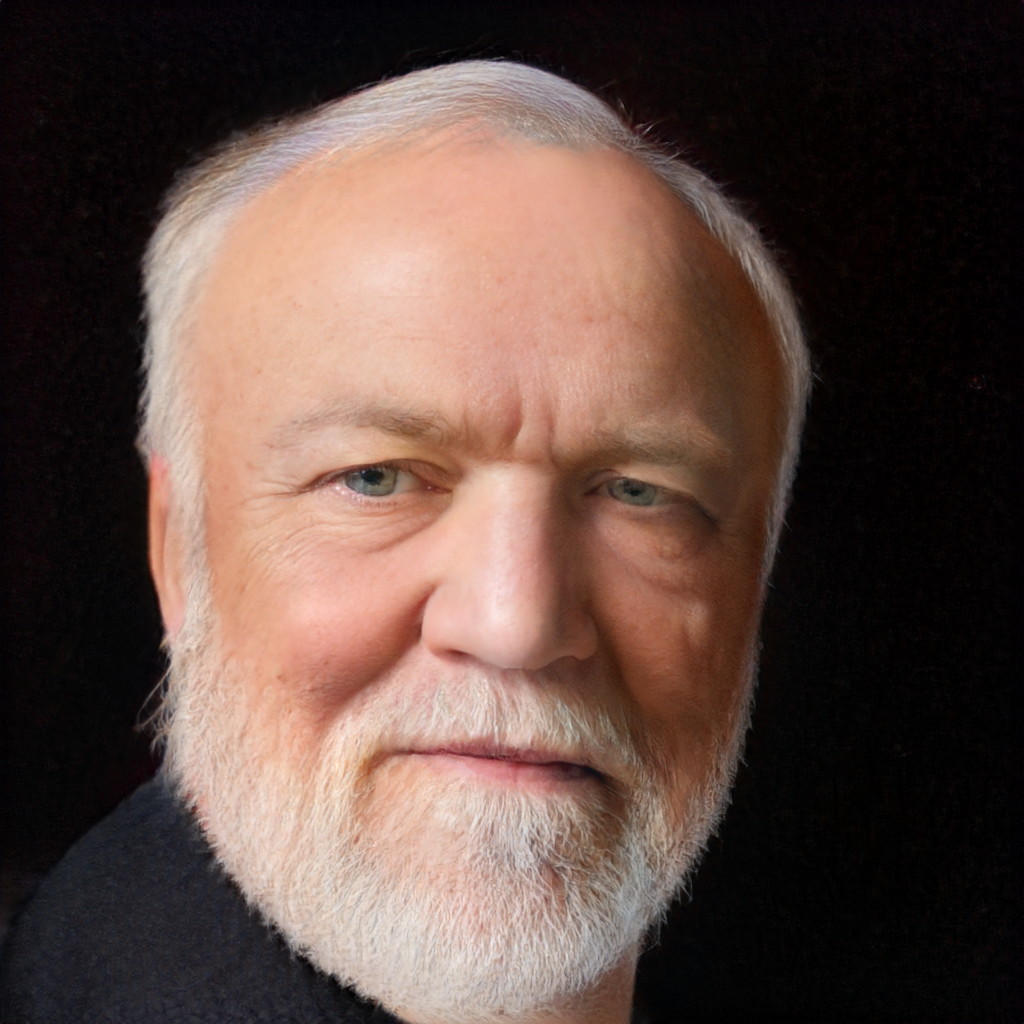} &  \includegraphics[width=\imW]{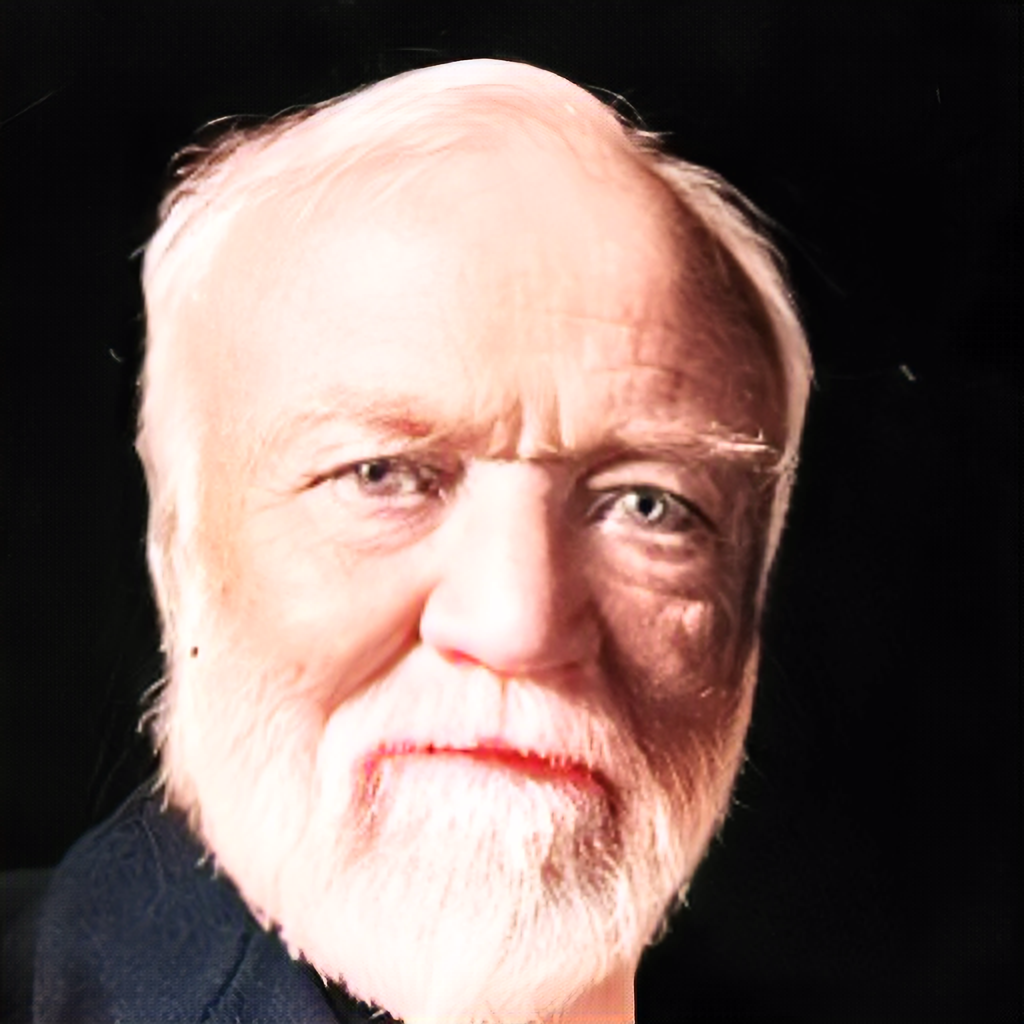} &
        \includegraphics[width=\imW]{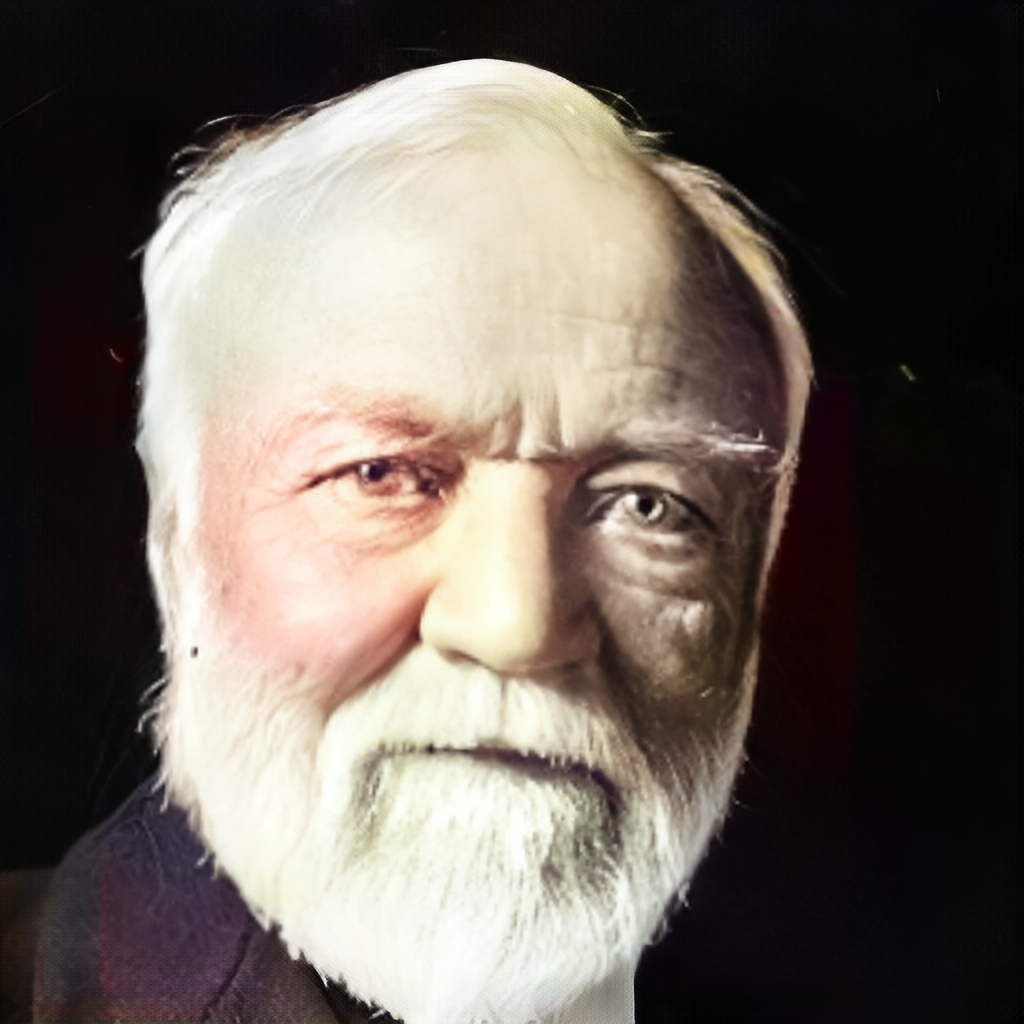} &
        \includegraphics[width=\imW]{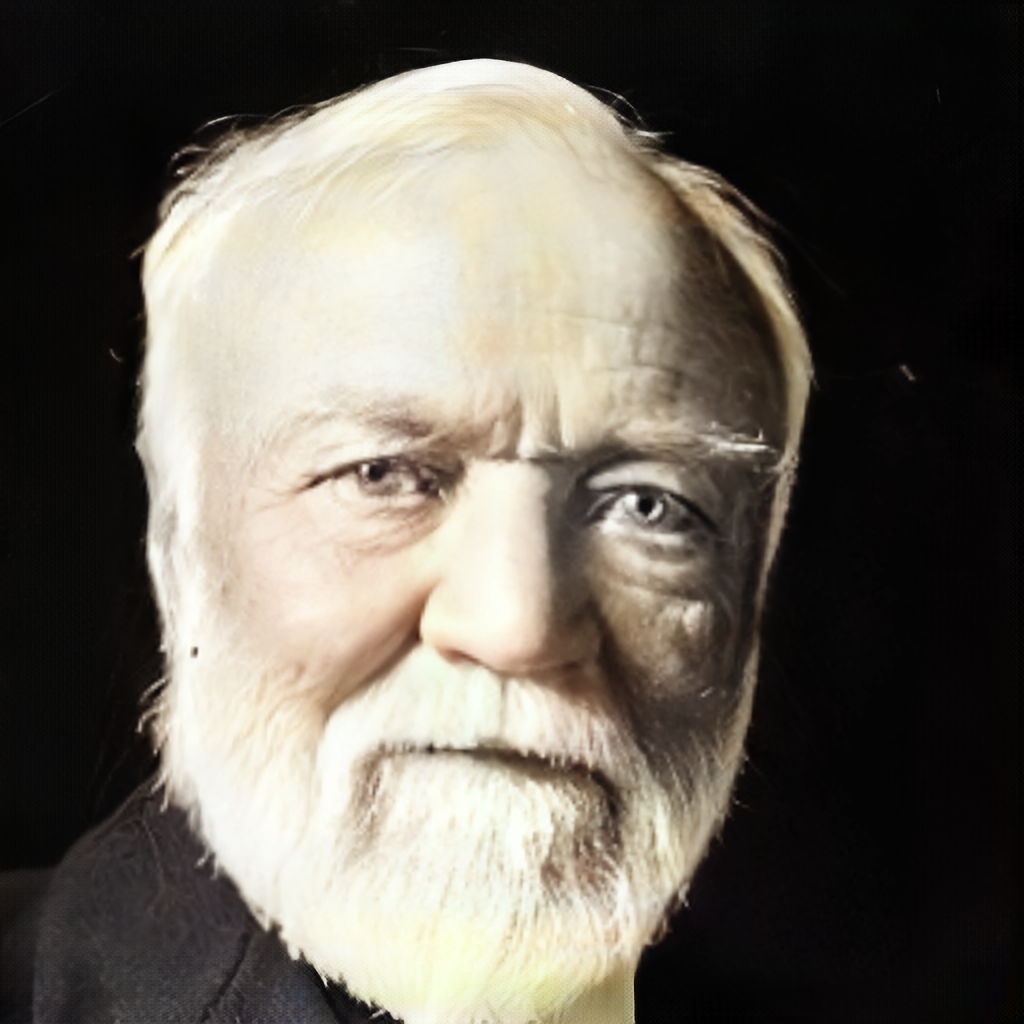} &
        \includegraphics[width=\imW]{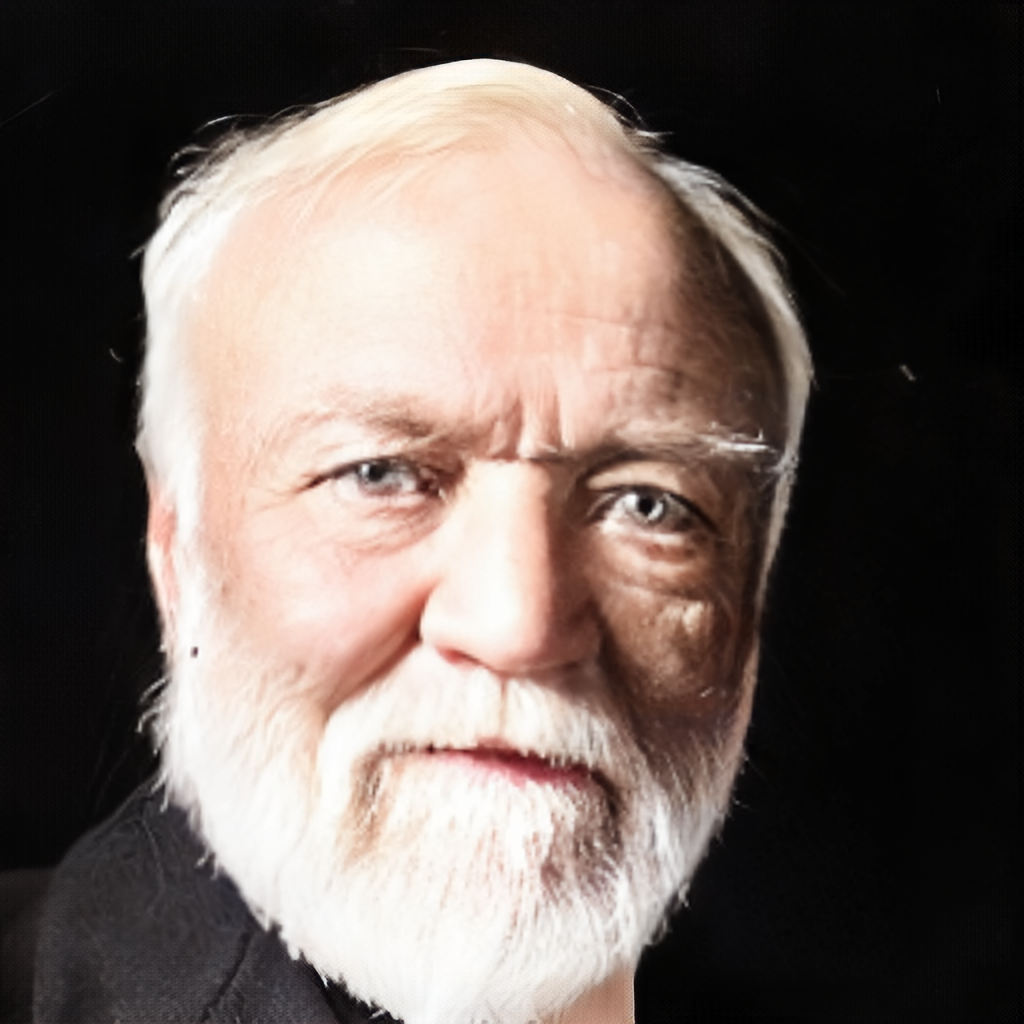}  \\

        \small{Input} & \small{Ours} & \small{DeOldify} & \small{InstColorization} & \small{Zhang} & \small{Zhang (FFHQ)}
        \vspace{-1.5em}
    \end{tabular}
    \caption{Comparisons of our approach to a pipeline built from published techniques for restoring, colorizing, and super-resolving old photos. We evaluate against four baseline pipelines, each with a different colorization algorithm, detailed in Sec.~\ref{sec:experiments}. All of them fail to achieve the same realistic skin appearance and overall image quality as our approach. Input image: Andrew Carnegie (c. 1913) from Library of Congress.
    }
    \label{fig:colorization-comparison}
    \vspace{-1em}
\end{figure*}

\begin{figure*}[p]
    \vspace{-1em}
    \centering
    \setlength{\tabcolsep}{0pt}
    \def\imW{0.2\linewidth}
    \begin{tabular}{p{\imW}p{\imW}p{\imW}p{\imW}p{\imW}}
    \multicolumn{5}{c}{\includegraphics[width=0.975\linewidth]{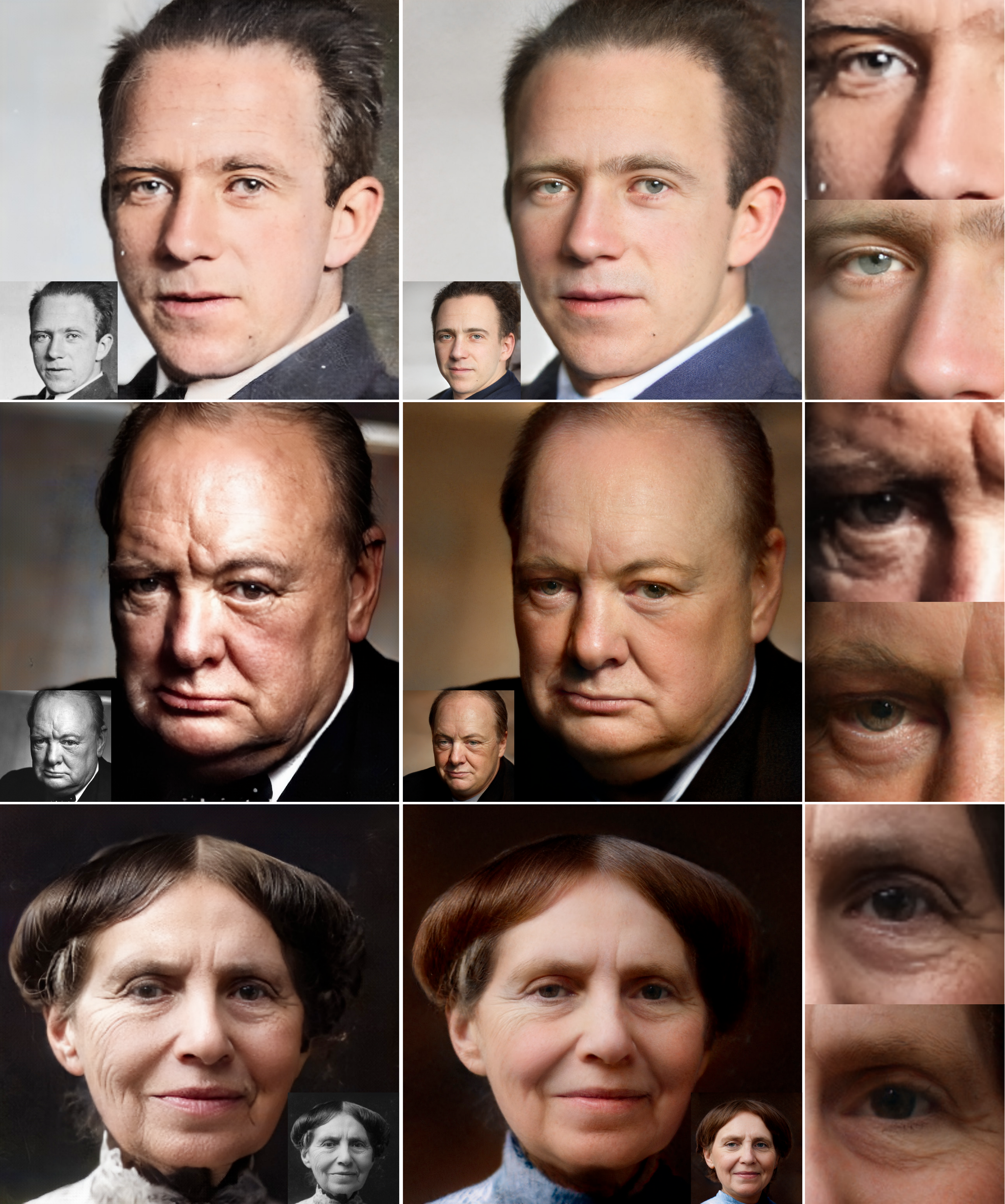}} \\
        \multicolumn{2}{c}{Zhang (FFHQ)} & \multicolumn{2}{c}{Ours} & \multicolumn{1}{c}{Zoom-in} \\
        & & & &
    \end{tabular}
    \vspace{-2.7em}
    \caption{Visual comparison with the top-performing baseline, \emph{Zhang (FFHQ)}. The input and sibling images are shown as insets in the first and second columns, respectively. Results are best evaluated at $1024\times 1024$.
    From top to bottom:  Werner Heisenberg (1933) by BArch, Bild 183-R57262 / Unknown author / CC-BY-SA 3.0, Winston Churchill (1941) by Library and Archives Canada / flickr, and Clara Barton (c. 1904) by Library of Congress.
    }
    \label{fig:comparison-highres}
\end{figure*}

\mypar{Historical Wiki Face Dataset.} 
We collect a list of names from the ``Significant people'' section of the Wikipedia pages for ``19th Century'' and ``20th Century'', and crawl down the main \emph{page images}~\cite{wiki-pageimages} on their Wikipedia pages. To filter out images unsuitable for our task, we remove the image if it belongs to any of the following categories:
1) The image is not a photo (e.g., a painting) or is in color. 2) The face is too small, i.e., less than 130 pixels. 3) The head is not fully visible. 4) Hands touch or occlude the face. 5) Face detection~\cite{dlib09} fails. %
6) The photo is heavily retouched (e.g., manually colorized daguerreotypes). 7) The subject is politically controversial.
The result of this filtering is 224 photos of unique historical figures dating from 19th century to 20th century, including people like Abraham Lincoln, Marie Curie, Winston Churchill, and Franz Kafka (e.g., Figs~\ref{fig:teaser},~\ref{fig:comparison-highres}). We cropped out the head region using the face alignment method of Karras et al.~\shortcite{karras2019style} and resized them to have a maximum resolution of $1024 {\times} 1024$ (average $638{\times}638$, min $133{\times }133$).
Our test set covers a wide range of image quality, head poses, genders, ethnic groups, and historical fashion styles that often drastically differ from modern ones.
Subjects in the early 19th century, for instance, often have big curly mustaches, long beards, and shaggy hair. Additionally, many antique accessories are uncommon in modern imagery, such as crowns, stand-up collars, and pince-nez. Fig.~\ref{fig:teaser},~\ref{fig:spectral-sensitivity},~\ref{fig:camera-response},~\ref{fig:colorization-comparison},~\ref{fig:failure} all include samples from this test set.

\paragraph{Experiment Setup.} In our experiments we manually select the input blur kernel. For film model, we use the blue-sensitive model for photos before 1873, manually select between blue-sensitive and orthochromatic for images from 1873 to 1906 and among all models for photos taken afterwards.

\begin{table}[t]
    \centering
    \small
    \caption{\update{Quantitative comparisons between our approach and baseline pipelines composed of restoration, colorization and super-resolution methods. The NIQE score of the input images is also reported as a reference.}\\
    }
    \vspace{-2em}
    \label{tab:quant-comparison}
    \resizebox{\linewidth}{!}{
    \renewcommand{\arraystretch}{1.1}
    \begin{tabular}{lrrrrrr}
        \toprule
            &Input & DeOldify & Zhang & InstColorization & Zhang (FFHQ) & Ours \\
        \midrule
            NIQE $\downarrow$  &     7.08 & 5.54 & 5.52    & 5.49   & 5.47   & \bf{4.55} \\
        \bottomrule
    \end{tabular}
    
    }
    \vspace{-1.5em}
\end{table}

\begin{figure}[t]
    \centering
    \vspace{5pt}
    \includegraphics[width=\linewidth]{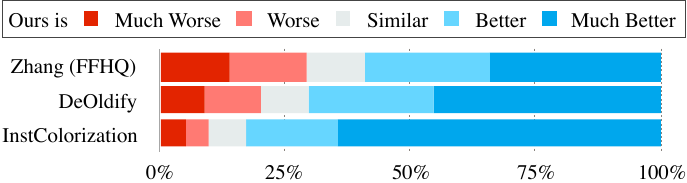}
    \vspace{-2em}
    \caption{User ratings of our approach compared to baselines composed of state-of-the-art restoration, colorization and super-resolution methods. %
    }
    \vspace{-1.5em}
    \label{fig:user-study}
\end{figure}

\newcolumntype{Y}{>{\centering\arraybackslash}X}
\begin{figure}[t]
    \centering
    \small
    \begin{tabularx}{0.95\linewidth}{@{}YY}
        \multicolumn{2}{l}{\hspace{-1em}\includegraphics[width=\linewidth]{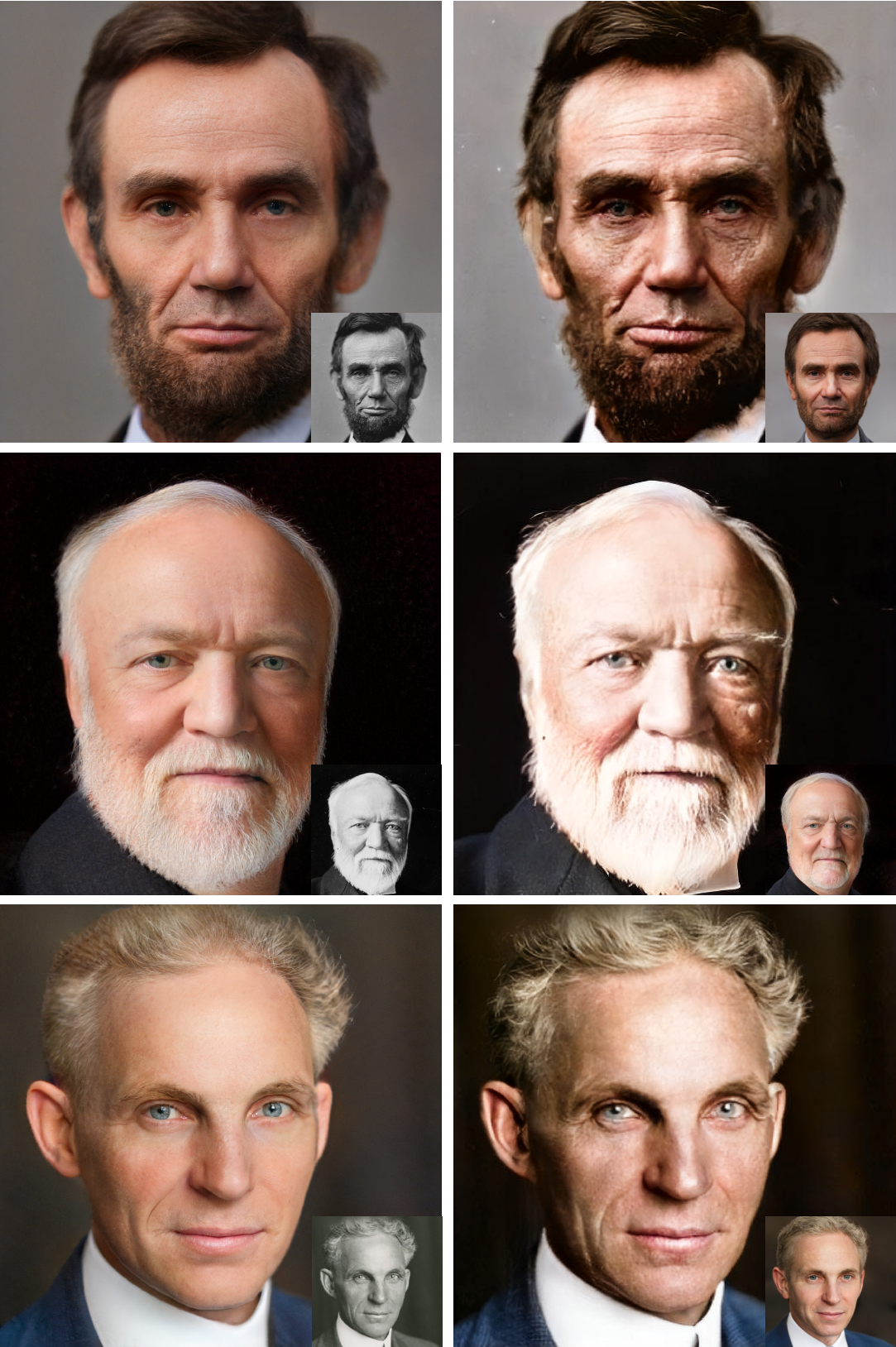}}\\
        Ours & Exemplar-based Colorization
    \end{tabularx}
    \vspace{-1.5em}
        
    \caption{\update{Visual comparisons with the exemplar-based colorization technique by He et al.~\shortcite{he2018deep} using our sibling images as the exemplars. The input and sibling images are shown as insets in the the first and second columns, respectively. Input images (top to bottom): Abraham Lincoln (1863) from Mead Art Museum, Andrew Carnegie (1913) from Library of Congress, Henry Ford (1863 - 1947) from the Collections of the Henry Ford.}}
    \label{fig:exemplar}
    \vspace{-2.5em}
\end{figure}

\newcolumntype{Y}{>{\centering\arraybackslash}X}

There is no published baseline method that performs the full
complement of image restoration operators needed for antique photos, i.e., noise+blur removal, contrast adjustment, colorization, and super-resolution. 
We therefore compare our approach to sequentially applying state-of-the-art methods for each of these tasks. As a first step, we apply Wan et al.~\shortcite{wan2020}, which was specifically designed to remove noise and artifacts in antique portrait photos at resolution $512{\times} 512$. We tried restoring at $1024 {\times} 1024$ but found the method produced blurrier and more noisy results compared to apply at $512{\times} 512$ followed by a separate super-resolution technique (detailed later). As a second step, we colorize the image. We evaluated several colorization techniques including DeOldify~\cite{antic2019deoldify}, InstColorization~\cite{Su-CVPR-2020}, Zhang et al.~\shortcite{zhang2017real}. All of these methods are designed for generic scenes and perform worse on antique portraits. We therefore retrained Zhang's colorization network using the FFHQ dataset of face images~\cite{karras2019style}, denoted \emph{Zhang (FFHQ)}. We also augmented this training dataset by applying random Gaussian blur and noise to make their method more robust to antique imagery. As a final step, we use the SRFBN~\cite{li2019feedback} (BI model) to super-resolve ($2\times$) the colorized output to our target resolution of $1024{\times}1024$. For simplicity, we use the name of the colorization method to refer to the full pipeline baseline that is composed of Wan et al.~\shortcite{wan2020}, one of the four colorization methods, and SRFBN~\cite{li2019feedback}.

\paragraph{Qualitative Evaluation.}
Figure~\ref{fig:colorization-comparison} shows a comparison between our method and the pipelines described above, including each of the four colorization methods.
Our method outperforms the other baselines and produces much more photo-realistic skin texture and colors. 
Figure~\ref{fig:comparison-highres} shows a higher-resolution comparison between our method and our strongest baseline, \emph{Zhang (FFHQ)}. None of the baseline methods are able to reproduce realistic skin appearance and sharp image details as well as ours. %

\paragraph{Quantitative Evaluation.} \update{Table~\ref{tab:quant-comparison} quantitatively compares our method with the baseline pipelines on their abilities to resolve degradations in the luna component, such as denoising, deblurring, enhancing contrast and details. We extract the \emph{Y'} channel from the \emph{Y'CrCb} color space for all the color images, and measure the quality of the restored results using the no-reference metric \emph{NIQE}~\cite{mittal2012making}. We also report the NIQE score for the input images as a reference. Our results is shown to produce more details and outperform all baseline methods in NIQE.}

\paragraph{User Study.} We conducted a user study over the \emph{Historical Wiki Face Dataset}. 
39 users participated in the study, where each participant was presented a random set of 56 or 112 pairwise comparisons of our result and one of the three top-performing baseline pipelines (\emph{Zhang (FFHQ)}, \emph{DeOldify} and \emph{InstColorization}). The input image was also presented.
The participants were asked to choose one of the two results and rated among ``significantly better'', ``slightly better'', or ``similar'', when asked ``Which of the two images is a higher-quality portrait?'' We obtained answers for all 224 images in our dataset and all three baseline methods. The results of this study are presented in Fig.~\ref{fig:user-study}. Our approach is consistently and significantly preferred over \emph{InstColorization} and \emph{DeOldify}. Even compared with the strongest baseline, $58.9\%$ of the time participants perceived our results to be better compared to $29.4\%$ for \emph{Zhang (FFHQ)}. $11.7\%$ expressed no strong preference.

\paragraph{Antique vs. Modern Photographic Processes.} \xuan{In Fig.~\ref{fig:old_vs_new}, we use the gelatin dry plate photos as input, restore the images with the strongest baseline, ~\emph{Zhang (FFHQ)}, and compare with our method using the blue-sensitive model. In the top row,  pimples and freckles are dramatically exaggerated, and skin appears more weathered and specular in the blue-sensitive gelatin dry plate. The results by \emph{Zhang (FFHQ)} further exaggerate these features. While being slightly smooth, our results restore a more natural skin texture and specularity that better resembles the DSLR reference. Similarly, in the bottom row, the baseline exaggerates skin defects in the forehead and adds sharp specular effects that are reduced in our result.
}
\paragraph{Comparison with Exemplar-based Colorization.} Fig.~\ref{fig:exemplar} compares our method with prior exemplar-based colorization method~\cite{he2018deep}. 
Similar to other baselines, We apply the same restoration~\cite{wan2020} and super-resolution~\cite{li2019feedback} method but adopt the colorization method by He et al.~\shortcite{he2018deep} using our sibling as the exemplar.
Our sibling images as exemplars improves the overall color and tone, but the method still fails to remove artifacts in the luminance channel, nor does it account for the unique spectral response properties of antique negatives. As a result, our method produces more detailed and realistic results.

\section{Limitations and Future Work}
\label{sec:limitations}

\begin{figure}[t]
    \centering
    \includegraphics[width=0.95\linewidth]{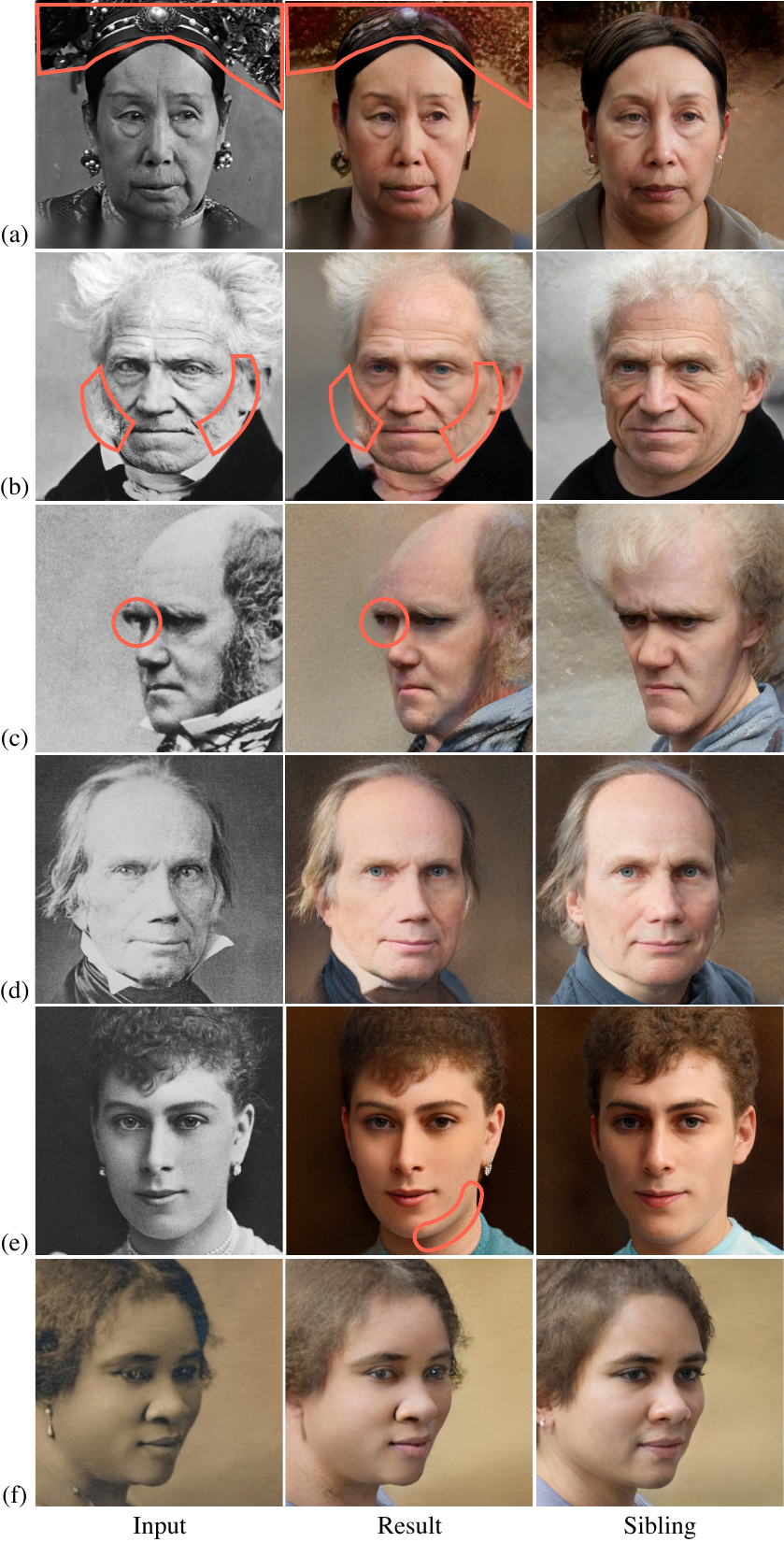}
    \vspace{-1em}
    \caption{
        \small{
            Our approach can struggle with features that are not well represented in \stylegantwo, such as uncommon accessories, clothing, (facial) hairstyles, and extreme head poses (a-c). Extremely poor image quality or severely compressed intensity gamuts may also limit the quality of the result (d). The sibling image may also present a different gender or ethnicity than the input (e-f). Input images (top to bottom): Cixi, Empress Dowager of China (1835 - 1908), FSA A.13, Freer Gallery of Art and Arthur M. Sackler Gallery Archives, Arthur Schopenhauer (1859) from Goethe-Universitat Frankfurt am Main, Charles Darwin (c. 1854) by GrrlScientist/flickr, Henry Clay (1848) by Jim Surkamp/flickr, Mary of Teck (1867 - 1953) from Royal Collection Trust/\copyright Her Majesty Queen Elizabeth II 2021, and Madam C. J. Walker (c. 1914) from Smithsonian Institution, National Museum of American History.
        }
    }
    \label{fig:failure}
    \vspace{-1em}
\end{figure}

\xuan{
As illustrated in Fig.~\ref{fig:failure}, our method does not work equally well on all images, and 
inherits biases from the \stylegantwo{} image generator.
Historical hairstyles, accessories, and clothing that are very different from anything present in the \stylegantwo{} training set are not reproduced well. Extreme head poses are also rare in the training data and harder to restore. In these cases, it is challenging to create a \stylegantwo{} sibling with such uncommon features (Fig.~\ref{fig:failure}a-d),
leading to inferior results.  %
In some cases the sibling presents a different gender or ethnicity, which affects the synthesized result (Fig.~\ref{fig:failure}). As shown by Salminen et al. ~\shortcite{salminen2020analyzing}, StyleGAN's generated images have strong biases towards younger and White people: $72.6\%$ of generated images represent White, $13.8\%$ Asian, $10.1\%$ Black, and $3.4\%$ Indian people. 
These biases in StyleGAN in turn lead to biases of our method towards predicting lighter skin tones for some inputs.
Note that the brightness and contrast of the input image can also affect the predicted skin color.
Addressing these gender and ethnicity shift problems is an important topic of future research. %
We believe this problem can be addressed by training \stylegantwo{} with datasets of balanced race, gender, and age, such as the \emph{FairFace}~\cite{karkkainenfairface} dataset, which in turn can improve our method with a more balanced sibling encoder.
}

\xuan{Another limitation of our method is inaccurately predicting skin texture from images with compressed intensity gamuts (Fig.~\ref{fig:failure}).
While camera response fitting alleviates poor exposure and contrast from the antique photos to some extent, handling more challenging cases requires hallucinating higher dynamic range.
Our performance also degrades on images with severe noise.
}

\xuan{
In some cases, our method alters the shape of certain facial features such as the eyes (Kafka's right eye in Fig.~\ref{fig:comparison-highres}), wrinkles (Lincoln's forehead wrinkles in Fig.~\ref{fig:teaser}), glasses (glasses of Gandhi in Fig.~\ref{fig:teaser}), etc. This is because we only optimize global $\mathcal{W}+$ style codes with a strong noise regularizer, making it hard to preserve local image details. Integrating local features from the input image with \stylegantwo{} face prior is another direction for future work.}

\xuan{
Recovering the correct color for skin, eyes, or clothes, is challenging (Fig.~\ref{fig:old_vs_new}) as many possible colors can correspond to the same degraded photo. One avenue of future work is to predict the distribution of likely color outputs, or guide the color prediction using references from paintings and textural descriptions.
}

\section{Conclusion}

We introduced \emph{time-travel rephotography}, an image synthesis technique that simulates rephotographing famous subjects from the past using a modern high-resolution camera based on a black-and-white reference photo. Our basic approach is to project this reference image into the space of modern high-resolution images represented by the \stylegantwo{} generative model~\cite{karras2020analyzing}. This is accomplished through a constrained optimization over latent style codes that is guided by a novel reconstruction procedure that simulates the unique properties of old film and cameras. We also introduce our {\em sibling} encoders that generate an image to recover colors and local spatial details in the result. Improving on applying a sequence of state-of-the-art techniques for image restoration, colorization, and super-resolution, our unified approach is able to render strikingly realistic and immediately recognizable images of historical figures.
\begin{acks}
\xuan{We thank Roy Or-El, Aleksander Holynski and Keunhong Park for insightful advice. 
This work was supported by the UW Reality Lab, Amazon, Facebook, Futurewei, and Google.}
\end{acks}

\bibliographystyle{ACM-Reference-Format}
\bibliography{egbib}

\end{document}